\pdfoutput=1

\documentclass[11pt]{article}

\usepackage[preprint]{acl}

\usepackage{times}
\usepackage{latexsym}

\usepackage[T1]{fontenc}

\usepackage[utf8]{inputenc}

\usepackage{microtype}

\usepackage{inconsolata}

\usepackage{graphicx}

\usepackage[T1]{fontenc}
\usepackage{times}
\usepackage{latexsym}
\usepackage{hyperref}
\usepackage{inconsolata}
\usepackage{url}
\usepackage{amsmath}
\usepackage{amsthm}
\usepackage{amsfonts}
\theoremstyle{definition}

\usepackage{graphicx}
\usepackage{subcaption}
\usepackage{booktabs}
\usepackage{multirow}
\usepackage{makecell}
\usepackage{wrapfig}
\usepackage{enumitem}
\usepackage{comment}
\usepackage{blindtext}
\usepackage{xcolor}
\usepackage{svg}
\usepackage{xspace}
\usepackage{adjustbox}

\renewcommand{\cite}{\citep}

\usepackage{listings}
\lstdefinelanguage{prompt}{
    frame=l,
    framerule=3pt,
    framesep=8pt,
    postbreak=\mbox{$\hookrightarrow$\medspace},
    basicstyle=\scriptsize\ttfamily,
    commentstyle=\color{cyan},
    morecomment=[l]{//},
    moredelim=[is][\color{red}\bfseries]{<<<}{>>>},
    moredelim=[is][\color{magenta}\bfseries]{[[[}{]]]},
    moredelim=[is][\color{orange}\bfseries]{===}{===},
    moredelim=[is][\color{olive}\bfseries]{|||}{|||},
}
\lstdefinelanguage{ioexample}{
    frame=shadowbox,
    rulesepcolor=\color{gray},
    framerule=0.5mm,
    rulesep=2mm,
    basicstyle=\small\normalfont,
    commentstyle=\color{cyan},
    morecomment=[l]{//},
    moredelim=[is][\color{red}\bfseries]{<<<}{>>>},
    moredelim=[is][\color{magenta}\bfseries]{[[[}{]]]},
    moredelim=[is][\color{orange}\bfseries]{===}{===},
    moredelim=[is][\color{olive}\bfseries]{|||}{|||},
    moredelim=[is][\bf]{:::}{:::},
    moredelim=[is][\it]{---}{---},
    moredelim=[is][\tt]{+++}{+++},
}

\usepackage[T1]{fontenc}

\usepackage[utf8]{inputenc}

\usepackage{microtype}

\usepackage{inconsolata}

\usepackage[]{todonotes}

\usepackage{amsthm}

\newcounter{rqsection}

\renewcommand{\therqsection}{RQ \arabic{rqsection}}

\newcommand{\rqsection}[1]{
  \refstepcounter{rqsection} %
  \medskip
  \noindent\textbf{\therqsection: \emph{#1}} %
}

\newcommand{\datasetname}{\textsc{PhysiCo }}
\newcommand{\datasetnamens}{\textsc{PhysiCo}}

\newcommand{\coredataset}{\datasetnamens-\textsc{Core }}

\newcommand{\coredatasetns}{\datasetnamens-\textsc{Core}}
\newcommand{\harddatasetns}{\datasetnamens-\textsc{Associative}}

\title{\emph{The Stochastic Parrot on LLM's Shoulder:}\\ A Summative Assessment of Physical Concept Understanding}

\newcommand{\authorsep}{\quad}

\author{%
Mo Yu$^1$\thanks{Equal contribution.}\authorsep
Lemao Liu$^1$\footnotemark[1]\authorsep
Junjie Wu$^2$\footnotemark[1]\authorsep
Tsz Ting Chung$^2$\footnotemark[1]\authorsep
Shunchi Zhang$^3$\footnotemark[1]\authorsep
\\
\bfseries
Jiangnan Li$^1$\authorsep
Dit-Yan Yeung$^2$\authorsep
Jie Zhou$^1$\authorsep
\\
\textsuperscript{1}WeChat AI, Tencent\authorsep
\textsuperscript{2}HKUST\authorsep
\textsuperscript{3}JHU\\
\texttt{moyumyu@global.tencent.com}\authorsep
\texttt{redmondliu@tencent.com}\\
\texttt{\{junjie.wu,ttchungac\}@connect.ust.hk}\authorsep
\texttt{szhan256@cs.jhu.edu}\\
{\hypersetup{urlcolor=magenta} \url{https://physico-benchmark.github.io}}
}

\begin{document}

\maketitle

\begin{abstract}
In a systematic way, we investigate a widely asked question: \emph{Do LLMs really understand what they say?}, which relates to the more familiar term \emph{Stochastic Parrot}.
To this end, we propose a summative assessment over a carefully designed physical concept understanding task, \datasetnamens.
Our task alleviates the memorization issue via the usage of grid-format inputs that abstractly describe physical phenomena.
The grids represents varying levels of understanding, from the core phenomenon, application examples to analogies to other abstract patterns in the grid world.
A comprehensive study on our task demonstrates: (1) state-of-the-art LLMs, including GPT-4o, o1 and Gemini 2.0 flash thinking, lag behind humans by $\sim$40\%; (2) the stochastic parrot phenomenon is present in LLMs, as they fail on our grid task but can describe and recognize the same concepts well in natural language;
(3) our task challenges the LLMs due to intrinsic difficulties rather than the unfamiliar grid format, as in-context learning and fine-tuning on same formatted data added little to their performance.
\end{abstract}

\section{Introduction}

Recent years have witnessed remarkable advancements in large language models (LLMs)~\cite{brown2020language,achiam2023gpt,team2023gemini}. 
Thanks to the substantial model capacity and massive training data, LLMs have achieved new state-of-the-arts on a variety of NLP tasks, even surpassing humans on some of them~\cite{min2023recent,chang2024survey}.
Nowadays the application of LLMs has become widespread, facilitating daily work and life, and profoundly influencing people's work and lifestyles~\cite{bommasani2021opportunities,peng2024large,demszky2023using}.

\begin{figure}[t!]
\centering
\includegraphics[width=0.94\columnwidth]{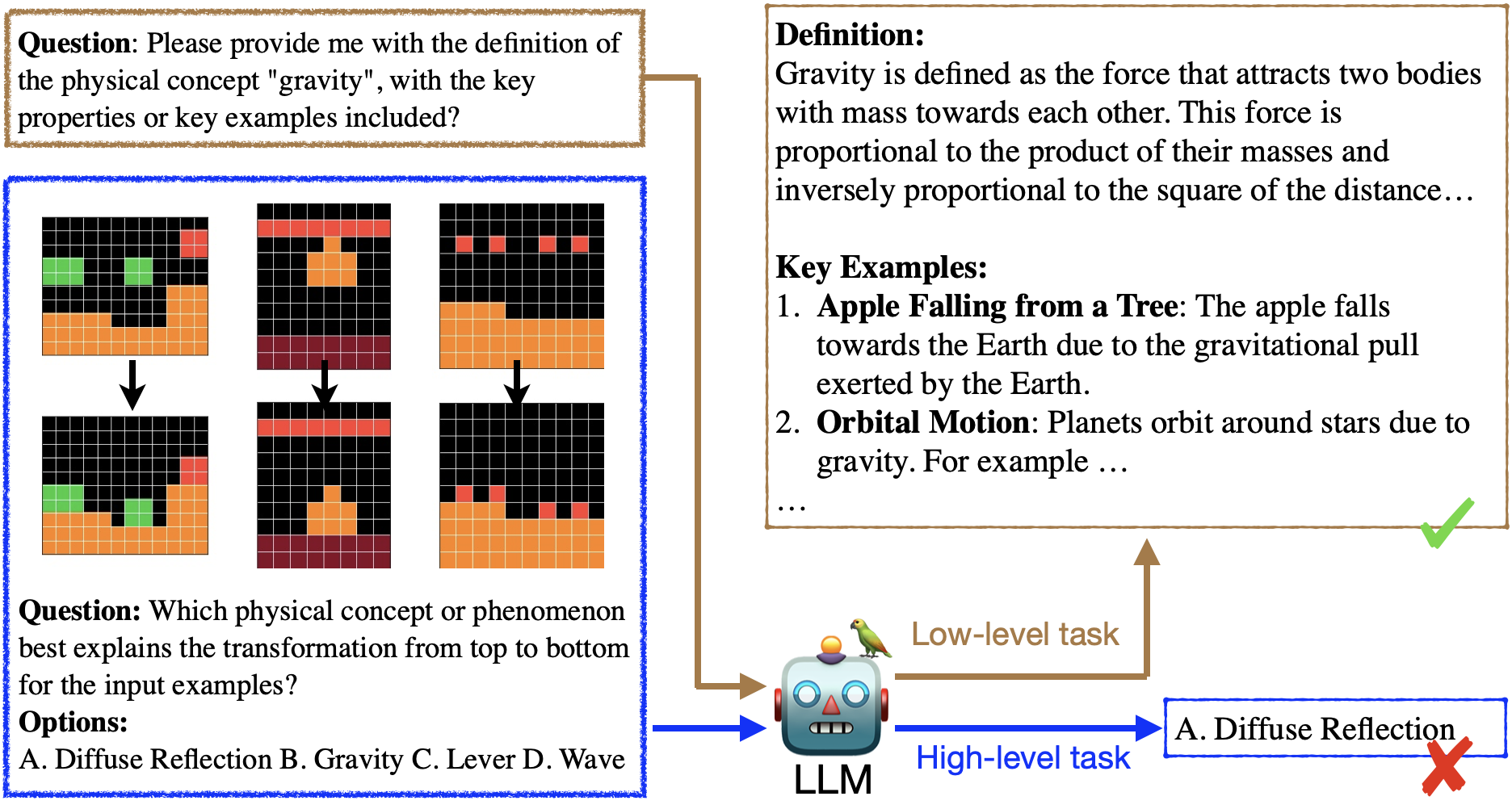}
\vspace{-0.1in}
\caption{Illustration of a ``Stochastic Parrot'' by our \datasetname task consisting of both \textcolor{brown}{low-level} and \textcolor{blue}{high-level} subtasks in parallel. For a concept {\em Gravity}, an LLM can generate its accurate description in natural language, but cannot interpret its grid-format illustration.}
\label{fig:teaser}
\vspace{-2ex}
\end{figure}

On the other hand, despite the great success of LLMs, many researchers argue that {\em LLMs may not really understand what they claim they do}~\cite{bender2020climbing,bender2021dangers,bommasani2021opportunities,mitchell2023debate} due to their strong memorization ability.
In particular, \citet{bender2021dangers} questioned whether LLMs are just \textit{Stochastic Parrots} that repeat words based on correlations without true understanding. 
This argument has been acknowledged by many research papers and dozens of them even include this term in their titles.\footnote{\url{https://scholar.google.com/scholar?hl=en\&q=llms+are+stochastic+parrot}.} Unfortunately, to our best knowledge, there are no quantitative experiments to verify the stochastic parrot phenomenon in LLMs. Existing studies indicate that LLMs may fail on one particular challenging task~\cite{chakrabarty-etal-2022-flute,shapira-etal-2023-well,hessel-etal-2023-androids,tong2024metaphor}, but they do not demonstrate that LLMs claimed to understand those tasks by providing a controlled and paired evidence. 

This paper aims to provide quantitative evidence to validate the argument of stochastic parrot in LLMs. To this end, from the perspective of educational and cognitive psychology, we first employ the approach of summative assessment~\cite{black1998assessment,black1998inside} to measure understanding in LLMs. Its key idea is to design various tasks that test different understanding levels regarding a specific concept.
Following the principle of Bloom's taxonomy~\cite{armstrong2010bloom,krathwohl2002revision}, we design tasks that reflect different levels of understanding.
Consequently, we develop \datasetnamens, a task designed to assess understanding of basic physical concepts from high school such as {\em Gravity}.
Our focus on physical concepts stems from both their fundamental relevance to important topics of world models and embodied systems~\cite{savva2019habitat,duan2022survey,xiang2024language}, and their rich denotations and connotations that enable effective design of summative assessment tasks.

Specifically, ~\datasetname includes two subtasks corresponding to two coarse levels of understanding in Bloom's taxonomy, as shown in Figure~\ref{fig:teaser}. One is the low-level understanding subtask in the natural language format, aimed at measuring the remembering (or memorization) ability of LLMs. The other involves the same concepts but in an abstract representation format inspired by~\cite{chollet2019measure}, which is designed to measure the high-level understanding beyond remembering of LLMs.

We conduct comprehensive experiments on \datasetname with representative open-source and commercial LLMs.\footnote{{Throughout this paper, LLM refers to either standard text-only LLMs or large multimodal models for simplicity.}} We obtain two key findings: (1) State-of-the-art LLMs perform perfectly on the low-level understanding subtask ($>$95\% in Accuracy) but lags behind humans by a large margin ($\sim$40\% in Accuracy) on the high-level subtask, which verifies the stochastic parrot phenomenon in LLMs.
(2) Further analysis shows that our high-level subtask challenges LLMs due to the intrinsic difficulty of deep understanding rather than the unfamiliar format. 

This paper makes the following contributions:
\begin{itemize}[noitemsep,nolistsep,leftmargin=*]
    \item We introduce a psychology-appealing approach (summative assessment) and a corresponding task \datasetname to measure the understanding of LLMs. 
    \item Based on \datasetnamens, we provide a quantitative experiment to successfully verify the stochastic parrot phenomenon in LLMs. 
    \item {As a by-product, our work presents a challenging comprehension task for existing text-only and multimodal LLMs, which establishes a substantial performance gap between humans and machines.}
\end{itemize}

\section{Measuring Concept Understanding via Summative Assessment}
\label{sec:towards}

It is intrinsically challenging to measure the extent to which LLMs {understand} a sentence or concept. Indeed, ~\citet{bender2020climbing} provide a definition of "understanding" from a linguistic perspective, but this definition depends on another abstract and unmeasurable term, ``\emph{meaning}''. 
Therefore, even with this definition, accurately measuring "understanding" remains elusive.

We approach the measurement of whether LLMs understand a concept from an educational and cognitive perspective, using \textbf{summative assessment}~\cite{black1998assessment,black1998inside,harlen1997assessment}.
Summative assessment is widely used by educators as an appealing strategy to evaluate students' understanding and knowledge acquisition in educational and cognitive psychology.
For example, when middle school physics teachers want to know whether a student truly understands the concept ``\emph{Gravity}'', they would design a series of questions specifically related to the concept of gravity to assess comprehension, \emph{e.g.}, the properties like inverse square law and examples like orbital motions. If a student struggles to answer many of these questions, the teacher may conclude that the student does not understand the concept well or has a poor grasp of it.

We extend the idea of summative assessment to evaluating the concept understanding of machines. Formally, assume $\mathcal{S}$ denotes an intelligent system and $\mathcal{C}$ is a specific concept.
To evaluate the extent how $\mathcal{S}$ understands the concept $\mathcal{C}$, our summative assessment includes the following two steps:
\begin{itemize}[noitemsep,nolistsep,leftmargin=*]
    \item \emph{Task design towards $\mathcal{C}$}: design several concept understanding tasks, each of which consists of several questions manually created towards understanding the concept $\mathcal{C}$.
    \item \emph{Evaluating $\mathcal{S}$}: 
    ask $\mathcal{S}$ to answer the questions from the tasks and calculate its accuracy. 
\end{itemize}

\paragraph{Requirements for Validity}
The success (validity) of the proposed evaluation approach highly depends on the task design~\cite{black1998assessment,black1998inside}. For example, if the questions are too easy, even a weak system could answer them correctly. This leads to an overestimation of the system's understanding capabilities, making the assessment ineffective. 
To ensure good validity, we adhere to the principles outlined in summative assessment~\cite{black1998assessment,black1998inside} for task design:
\begin{itemize}[noitemsep,nolistsep,leftmargin=*]
    \item \emph{Alignment with evaluating objectives}: the questions should be related to the targeted concept, and should measure the specific knowledge about the targeted concept. 
    \item \emph{Different difficulty levels}: the questions should be with different difficulty levels from easy to difficult level,  to ensure that the evaluation results have distinctiveness for different systems.
    \item \emph{Variety}: the questions should reflect various understanding aspects of the targeted concept; addressing both its denotation and connotation.
    \item \emph{Simplicity}: while not mandatory, a simpler benchmark for humans can more effectively highlight the issue faced by current models, i.e., the stochastic parrot effect in LLMs.
\end{itemize}

\section{Task Design and Dataset Construction}

\subsection{Task Design Principle}
We borrow the idea of Bloom's taxonomy~\cite{krathwohl2002revision,armstrong2010bloom} from education research to fulfill the requirements for task design in Section~\ref{sec:towards}, so as to ensure the assessment validity.
Bloom's taxonomy offers an ideal principle to these requirements with an ordering of six cognitive skills (from low to high level) for knowledge understanding: \emph{Remembering, Understanding, Applying, Analyzing, Evaluating and Creating}.

Generally, it is nontrivial to strictly follow this principle since there is no clear boundary among the last four skills of understanding. 
As a result, we group the last four high-level skills into one and consider the following two levels of understanding:
\begin{itemize}[noitemsep,nolistsep,leftmargin=*]
    \item \emph{Low-level Understanding}: covering the two lowest-level skills in Bloom's taxonomy, \emph{i.e.}, retrieving relevant knowledge from long-term memory and rephrasing in one's own words.
    \item \emph{High-level Understanding}: covering the aspects for understanding the knowledge beyond memorization. As shown by the examples in Section \ref{ssec:high-level}, our tasks directly correspond to a spectrum from the understanding level of applying to the level of analyzing in Bloom's taxonomy, 
    \emph{e.g.}, \emph{applying} the knowledge to explain a physical phenomenon, \emph{analyzing} a concrete property of a concept in a generalized and abstract manner,\footnote{For example, the flow of electric current can be abstracted as \emph{moving} from high potential to low potential.}. %
\end{itemize}
Based on these two levels, we design the following \datasetname task for summative assessment.

\subsection{Our \datasetname Task}

\datasetname is essentially a physical concept understanding task, which primarily targets on 52 physical concepts or phenomena: {\em e.g., gravity, light reflection, acceleration, buoyancy, inertia, etc} (see Appendix~\ref{app:dataset_details} for the full list).
Our focus on physical concepts is motivated by two main reasons:
1) understanding physical concepts is critical for intelligent systems to interact with the world, which is ultimate goal of embodied AI~\cite{savva2019habitat,duan2022survey,xiang2024language}; 
2) designing tasks centered around physical concepts allows us to more easily control different levels of understanding and ensure the diversity of each concept.

For each physical concept, \datasetname involves both low-level understanding subtasks and high-level subtasks, following our task design principles. %

\subsubsection{Low-level Understanding Subtasks}
\label{sec: low-level}

\paragraph{Physical Concept Selection (text)} 
First, to evaluate whether an LLM possesses the knowledge of our included concepts,
we design a task to recognize a concept from its corresponding Wikipedia definition.
Specifically, we manually masked the synonyms of the concept with placeholder {\small \texttt{[PHENOMENON]}}. Meanwhile, highly relevant entities were masked as {\small \texttt{[MASK]}} to alleviate shortcuts. For example, in the definition of \emph{Gravity}, the terms ``gravity'' and ``gravitation'' were masked as {\small \texttt{[PHENOMENON]}}, while ``Isaac Newton'' was masked as {\small \texttt{[MASK]}}. 
Details can be found in Appendix~\ref{app:rq1_details}.
We then present the LLMs with the same four choices as in our following high-level subtasks.

\paragraph{Physical Concept Selection (visual)} 
Second, we evaluate if the LLMs can recognize our concepts represented with real-life pictures.
To this end, we query our concepts on Google image search, and select the images that reflect the same core properties and examples annotated in our following high-level tasks.
This results in 100 examples.
We construct the same four-choice instances as above.

\paragraph{Physical Concept Generation}
Finally, we directly ask the LLMs to generate the description of a concept with its core properties and representative examples.
For instance, the concept \emph{Gravity} is described as ``\emph{a force that pulls objects with mass towards each other}'', followed by the example ``\emph{an apple falls to the ground}'' as shown in Figure~\ref{fig:teaser}.
We then evaluate the performance of LLMs by measuring the quality of the description and its coverage of knowledge required by our \datasetname and we employ both automatic and human metrics as presented in Section 5.2.
This provides a quantitative measure of the knowledge LLMs can recall in the context of our assessment.

\subsubsection{High-level Understanding Subtasks}
\label{ssec:high-level}
The low-level subtasks are depicted in natural language thus are likely to be remembered by the LLMs due to their extensive training data.
To assess whether the LLMs possess a deep understanding of the knowledge, we require the subtasks that can 1) represent the high-level understanding skills; 2) avoid the effects of memorization. 

The Abstraction and Reasoning Corpus (ARC)~\cite{chollet2019measure} provides a compelling way by using grids (or matrices) instead of texts to represent a concept. While the LLMs have seen matrices during pre-training, the data is less likely to be correlated to physical concepts. We hence adopt this idea to represent our subtask as abstract representations in the grid world that associate to the key properties of a physical concept. 

\paragraph{The \coredataset Set}

Our first subtask aims to cover the core properties or most representative examples/applications of the assessed concepts.
To ensure our set remains generally comprehensible to humans, we maintain a high school-level difficulty and selected 52 common physical concepts within the curriculum.
To enhance the diversity and richness, five annotators have labeled multiple core aspects of each concept. 
For example, the annotated core aspects of \emph{Gravity} include \emph{attraction between two bodies, motion on an inclined plane, objects falling to grounds} and \emph{orbital motions}.

For each aspect of a concept, the annotator is asked to draw several pairs of abstract grid representations. The aspect of the concept is guaranteed to be illustrated by the pair, such that it explains the transformation from the input to the output. For example, Figure~\ref{fig:teaser} forms a direct abstract visualization of the \emph{Gravity} concept from textbooks, \emph{i.e.}, \emph{apple falling from a tree}.
This results in 1,200 paired grid examples for the 52 concepts, which form 400 3-shot instances.

\begin{figure}[t!]
\centering
\includegraphics[width=0.8\columnwidth]{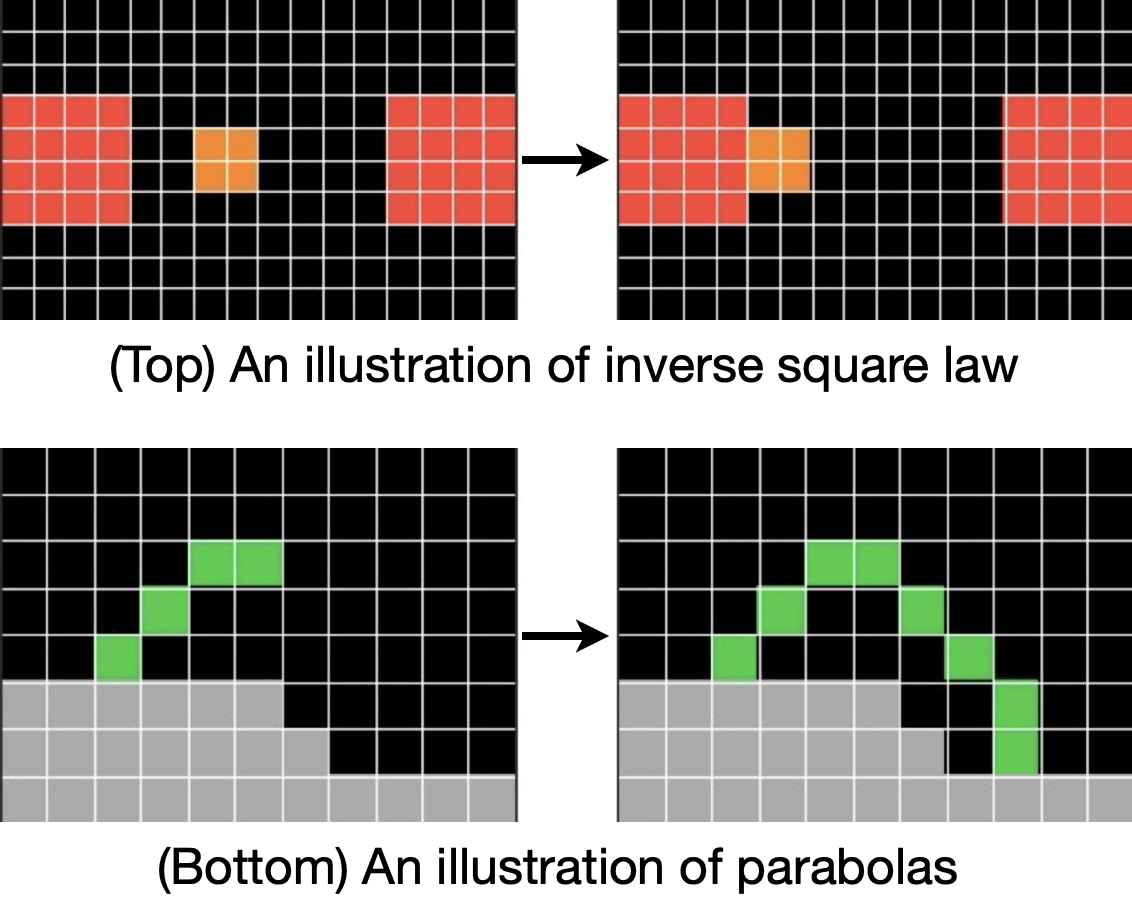}
\vspace{-0.1in}
\caption{Examples of input-output grids labeled as \emph{Gravity}, with increasing difficulty levels.}
\label{fig:level_examples}
\vspace{-0.25in}
\end{figure}

Figure~\ref{fig:level_examples} presents two examples from this subtask that delve deeper into the concept of \emph{Gravity}
compared to Figure~\ref{fig:teaser}.
The top example demonstrates an application of the \emph{inverse square law of gravity}.
The bottom one presents a parabola, linking the knowledge of \emph{gravity} to \emph{inertia}.
These examples demonstrate the difficulty of inferring their ground-truth labels solely by recalling the concept of \emph{Gravity} without high-level understanding skills.

\paragraph{The \harddatasetns\ Set}
Many instances in the original ARC dataset can be solved via association or analogy to physical concepts.
Therefore, as a second source of subtasks, we ask annotators to manually pick input-output grids from ARC that can evoke their associations to specific physical concepts and assign these concepts as ground-truth labels.
Different from \coredatasetns, we adopt an open-coding schema and allow the inclusion of new concepts during annotation.
The annotators have reviewed 500 ARC instances to filter out the required ones.
After cross-validation to ensure agreement, it results in a collection of 200 instances with physical concept labels.

This relabelling approach covers additional 15 physical concepts.
The resulted subtasks have each example represent an abstract aspect of a concept with possible distracting information. Consequently, the resulted task is more subjective hence more challenging than the \emph{\coredatasetns}Set.

\paragraph{Creation of Classification Tasks}

We create \emph{four-choice} tasks on the annotated data. Each instance consists of 3 unique grid pairs as input examples.
This results in 200 instances for \coredataset development set, 200 instances for \coredataset test set, and 200 instances for \textsc{Associative} respectively.
For each instance, we select three additional labels from our concept pool, along with the ground-truth label, as candidate options.
We manually avoid ambiguity during the negative sampling. For example, if \emph{Gravity} is the ground-truth, concepts like \emph{Magnet} will not be sampled.

\section{Overview of Our Studies}
\label{sec:overview}
In the following sections, we conduct a series of studies on our \datasetname tasks. Our studies are organized into six \emph{Research Questions (RQs)}, through which we aim to answer three \emph{Hypotheses} as shown in Figure~\ref{fig:research_overview}.
In summary, we propose to:
\begin{figure}[t!]
\centering
\vspace{-0.05in}
\resizebox{0.5\textwidth}{!}{\includegraphics[width=\columnwidth]{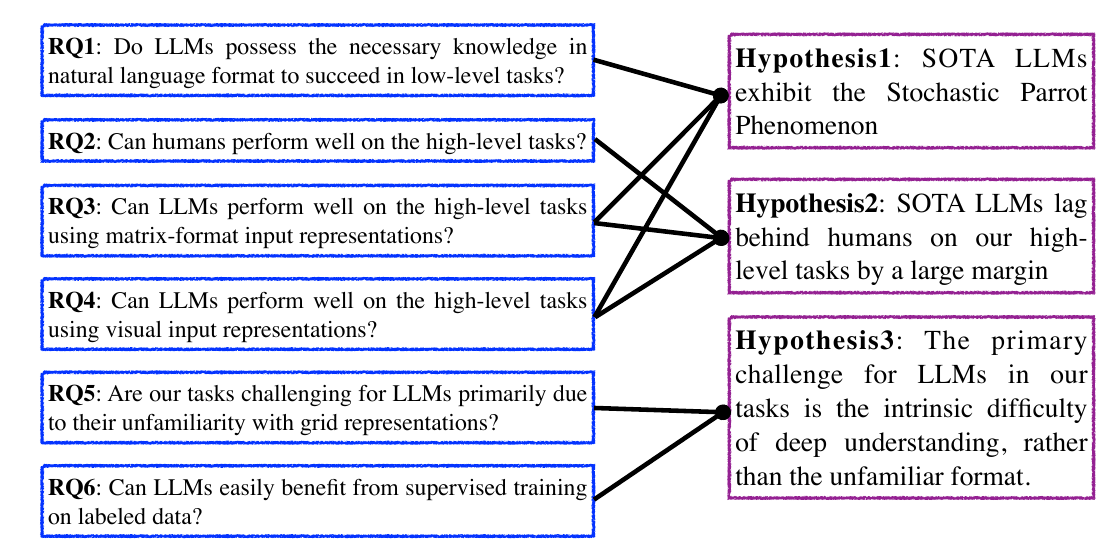}}
\vspace{-0.2in}
\caption{Overview of the research questions answered in our study and their relationships.}
\label{fig:research_overview}
\vspace{-0.2in}
\end{figure}

(1) Examine the quantitative disparity in LLMs' performances between low-level (\ref{rq:textual_input}) and high-level subtasks (\ref{rq:matrix_input}, \ref{rq:visual_input}). 
This aims to highlight \textbf{the existence of stochastic parrot phenomenon} in LLMs' understanding of physical concepts.

(2) Assess the performance gap between LLMs (\ref{rq:matrix_input}, \ref{rq:visual_input}) and humans on our high-level subtasks (\ref{rq:human_perf}). This aims to demonstrate that LLMs \textbf{fall significantly short of human understanding}.

(3) Investigate the shortcomings of in-context learning and supervised fine-tuning in improving LLMs on our high-level subtasks (\ref{rq:format_analysis}, \ref{rq:sup_training}). This aims to underscore the \textbf{intrinsic limitations} of SOTA LLMs in achieving deep understanding.

\paragraph{Experimented Models}
We use commercial LLMs, including GPT-3.5 ({\small \texttt{gpt-3.5-turbo-1106}}), GPT-4 and GPT-4v ({\small \texttt{gpt-4-turbo-2024-04-09}}), GPT-4o ({\small \texttt{gpt-4o-2024-05-13}});
and open-source LLMs, including Llama-3 ({\small \texttt{Llama-3-8B-Instruct}})~\cite{llama3} and Mistral ({\small \texttt{Mistral-7B-Instruct-v0.2}})~\cite{jiang2023mistral}, InternVL-Chat-V1-5 
~\cite{chen2023internvl,chen2024far}%
and LLaVA-NeXT-34B~\cite{liu2023improvedllava,liu2023llava}.
We use the default
inference configurations of the LLMs.
Considering the randomness, we run each experiment 3 times and compute the average and standard derivation.
We also experimented with the recent thinking models like o1.

\section{Validation on Low-Level Subtasks}

To illustrate the stochastic parrot phenomenon with \datasetnamens, a necessary condition is to ensure the LLMs can perform well on the low-level understanding subtasks, \emph{i.e.}, whether LLMs exhibit strong skills of \emph{recalling} and \emph{describing} the definitions, core properties and representative examples of the physical concepts in our tasks. That is:

\rqsection{Can LLMs perform well on low-level subtasks, i.e., understanding the definitions of physical concepts in natural language?}
\label{rq:textual_input}

To answer \ref{rq:textual_input}, we evaluate the LLMs' abilities to comprehend the definitions of these concepts and generate their descriptions and examples in natural language, as defined in Section~\ref{sec: low-level}.

\subsection{Concept Selection Subtask}
\paragraph{Settings} 
We provide the standard definition of a concept based on Wikipedia with its synonyms masked; then ask the LLMs to identify the concept, under the same four-choice setting throughout the experiments.
We evaluate the representative text-only LLMs and compute the accuracy. 

\begin{table}[t!]
    \small
    \centering
    \begin{tabular}{ccccc}
    \toprule
    \multirow{2}{*}{(a)}& \bf Mistral & \bf Llama-3 & \bf GPT-3.5 & \bf GPT-4 \\
     \cmidrule{2-5}
    & 81.0$_{\pm\text{1.3}}$& 88.5$_{\pm\text{0.7}}$& 97.3$_{\pm\text{0.3}}$ & 95.0$_{\pm\text{0.9}}$\\
    \bottomrule
    \toprule
    \multirow{2}{*}{(b)}& \bf InternVL & \bf LLaVA & \bf GPT-4v & \bf GPT-4o \\
     \cmidrule{2-5}
    & 66.3$_{\pm\text{7.7}}$ & 66.7$_{\pm\text{5.8}}$ & 93.7$_{\pm\text{0.9}}$ &93.7$_{\pm\text{0.5}}$\\
    \bottomrule
    \end{tabular}
    \vspace{-0.1in}
    \caption{Accuracy on the text-based (a) and visual-based (b) concept selection subtasks.}
    \label{tab:selection}
    \vspace{-0.2in}
\end{table}
\paragraph{Results} Table~\ref{tab:selection} shows that the GPT (both text-based and visual-based) models perform near perfect on 
recognition of our physical concepts from standard text-based definitions and from the real-life images.
Moreover, we observed that open-source models make more mistakes compared with the closed-source models due to the smaller model size. For the text-based models, both Mistral and Llama-3 are not as good as the closed-source models. Surprisingly, both InternVL and LLaVA are much worse than the open-source GPT models. One possible reason to this discrepancy is that our text-based concepts are from Wikipedia which is usually used as a part of the training data for open-source LLMs. In contrast, some of our selected images for those concepts may not be included in the training data of both InternVL and LLaVA which thereby can not memorize those visual instances.

\subsection{Concept Generation Subtask}
\paragraph{Settings}
We evaluate the descriptions LLMs generate for a concept. 
The evaluation of text generation is in general difficult. Moreover, in our scenario each concept have many different ground-truth examples in its description, thus existing automatic metrics such as BLEU~\cite{papineni2002bleu} and METEOR~\cite{banerjee2005meteor} are not capable of accurately measuring the quality. 
Therefore, we rely on mainly human evaluation for this subtask. We also propose an automatic metric via a self-play game for completeness in Appendix~\ref{app:self_play}.

\paragraph{Human evaluation metric} We ask the annotators to evaluate the quality of the generated descriptions. The evaluation uses binary scores: each description receives a score of 0 if it consists of any factual error on the concept itself
or any unfaithful examples,\footnote{For example, if the LLMs generated a wrong year in the description, it is not counted as incorrect physical knowledge.} 
and a score of 1 otherwise. %

\begin{table}[t!]
    \small
    \centering
    \begin{tabular}{cccc}
    \toprule
      \bf Mistral & \bf Llama-3 & \bf GPT-3.5 & \bf GPT-4 \\
     \midrule
    92.6& 100 &100 & 100\\
    \bottomrule
    \end{tabular}
    \vspace{-0.1in}
    \caption{Human evaluations on concept generation.}
    \vspace{-0.2in}
    \label{tab:generation}
\end{table}

\paragraph{Results}
The results of automatic and human evaluations are shown in Table~\ref{tab:generation}. 
According to human evaluation, there are no factual errors in the generated descriptions except for Mistral,
confirming that our selected concepts rely on basic and widely accepted knowledge.
Thought accurate, the open-source LLMs sometimes include correct but uncommon facts, \emph{e.g.,} listing single-slit diffraction as an example of \emph{Wave Interference}.
The additional self-play results in Appendix~\ref{app:self_play} further justify that all LLMs can accurately recognize the concepts from the descriptions they wrote by themselves.
Combining the conclusions, 
it shows the LLMs can generate correct and sufficient information.

\paragraph{Remark} We asked the annotators of our \textsc{Core} set to evaluate whether the core properties they annotated are covered by the LLMs' generated descriptions.
This corresponds to measuring the recall of the generated descriptions on core properties/examples of concepts from \coredatasetns. The recall rates for GPT-3.5 and GPT-4 are \emph{85.0} and \emph{90.0}, respectively. 
Of course, there are some exceptional examples from \coredatasetns\ missed in the descriptions. One example is that the LLMs fails to draw the connection between \emph{movable pulley} and the \emph{Lever} concept. Moreover, by manually checking these missed properties and examples, we found that most of them can be recalled if we query the LLMs in a second turn by prompting ``{\small \texttt{Any more core properties or examples?}}''. This confirms that the LLMs are \emph{aware of} and are \emph{able to recall} the core properties of concepts covered by the \coredatasetns, though some of them may not have the top conditional probabilities of generation.

\paragraph{Conclusion} LLMs understand the concepts covered by \datasetname in natural language format.
Notably, we find that the {properties and examples annotated in \coredatasetns\ are \emph{within the LLMs' knowledge} and are \emph{highly likely to pop up} when the corresponding physical concepts are queried}.

\section{Experiments on High-Level Subtasks}
This section answers the research questions regarding our high-level understanding subtasks.

\rqsection{Can Humans understand the abstract representations?}
\label{rq:human_perf}

First of all, we investigate the performance of humans who possess the knowledge required by our \datasetnamens.
For each instance in our \datasetnamens, we asked three independent annotators who were \emph{not} involved in our task design to perform the same classification task presented to the LLMs.
The annotators are provided with the same inputs used as prompts for the LLMs; and are permitted to consult GPT-4o for definitions of concepts they find unclear (mainly happens to the \textsc{Core}-Test set).
The results indicate that our tasks are largely solvable to people with a college-level education.
Specifically, on the \coredatasetns\ tasks, humans achieved an accuracy rate higher than 90\%.
The \harddatasetns\ tasks present greater challenges and subjectivity because the annotations are personalized based on the annotators' individual perspectives and experiences. Despite these challenges, humans can still achieve a notable average accuracy of 77.8\% in solving these tasks.

We conducted a detailed investigation into human performance on a subset of \harddatasetns. Participants were asked to annotate instances where they believed none of the four candidate answers adequately explained the inputs. The results revealed a 10.4\% rate of disagreement. On these disagreed-upon examples, human accuracy was 33.3\%, explaining a major factor for the human performance decline.

\paragraph{Conclusion} Our study demonstrates that {humans can perform the \datasetname tasks quite well}.

\rqsection{Can LLMs understand concepts in the abstract representations of the matrix format?}
\label{rq:matrix_input}

A straightforward solution for our \datasetname is to represent the grid-formatted examples as matrices. By representing the matrices with a token sequence, they can be integrated into an instruction prompt for text-based LLMs, following existing prompting methods for ARC tasks~\cite{acquaviva2022communicating,xu2023llms,wang2023hypothesis,wang2024speak}.
We use the prompt shown in Figure~\ref{fig:matrix_prompt_template} to query the answers from the evaluated LLMs.

\begin{table}[t!]
    \small
    \centering
    \resizebox{\columnwidth}{!}{
    \begin{tabular}{ll c||cc}
    \toprule
    & \multirow{2}{*}{\bf Models}  & \bf Dev  &\multicolumn{2}{c}{\bf Test}\\
    && \bf \textsc{Core}-Dev & \bf \textsc{Core}-Test& \bf \textsc{Assoc.}\\
    \midrule
    & Random & 25.0 & 25.0 & 25.0 \\
    \midrule
    \multirow{7}{*}{\rotatebox{90}{text-only}}& GPT-3.5  & 26.5$_{\pm\text{2.5}}$ &24.4$_{\pm\text{0.8}}$ & 30.0$_{\pm\text{2.5}}$ \\ %
    &GPT-4  & 41.3$_{\pm\text{1.3}}$ &28.2$_{\pm\text{2.3}}$ & 38.3$_{\pm\text{1.2}}$ \\ %
    &GPT-4o  & 34.0$_{\pm\text{2.9}}$ &31.3$_{\pm\text{2.9}}$ & 35.5$_{\pm\text{2.5}}$\\
    &\it o3-mini-high & 46.0$^*$ & \bf 46.5 & 42.5\\
    \cmidrule{2-5}
    &Mistral  & 21.5$_{\pm\text{0.3}}$ &26.0$_{\pm\text{1.4}}$& 23.2$_{\pm\text{0.4}}$\\
    &Llama-3  & 23.5$_{\pm\text{2.5}}$ &27.3$_{\pm\text{0.6}}$& 21.7$_{\pm\text{2.0}}$  \\
    &\it DeepSeek-R1  & 41.5 & 29.5 & \bf 55.0 \\
    \midrule
    \multirow{7}{*}{\rotatebox{90}{multi-modal}} & GPT-4v  & 34.2$_{\pm\text{1.6}}$ & 28.7$_{\pm\text{2.4}}$ & 32.0$_{\pm\text{1.5}}$ \\
    &GPT-4o  & 52.3$_{\pm\text{0.8}}$ & 45.2$_{\pm\text{2.3}}$ & 36.5$_{\pm\text{0.4}}$ \\ %
    &\quad +CoT  & 46.0$_{\pm\text{2.5}}$ &43.5$_{\pm\text{0.8}}$ &  39.5$_{\pm\text{1.1}}$\\
    &\it o1 & \bf {53.0} & 42.5 & 34.5 \\
    &\it Gemini2 FTE & 49.8$_{\pm\text{0.8}}$& 43.2$_{\pm\text{2.0}}$ & 36.8$_{\pm\text{3.1}}$ \\
    \cmidrule{2-5}
    &InternVL   & 26.3$_{\pm\text{1.6}}$ & 26.9$_{\pm\text{4.1}}$& 24.8$_{\pm\text{1.3}}$\\
    &LLaVA  & 26.2$_{\pm\text{1.1}}$  & 28.5$_{\pm\text{1.5}}$&24.7$_{\pm\text{3.2}}$ \\
    \midrule
    &Humans& 92.0$_{\pm\text{4.3}}$ & {89.5$_{\pm\text{5.1}}$} & 77.8$_{\pm\text{6.3}}$ \\
    \bottomrule
    \end{tabular}}
    \caption{Performance of different text-only and multi-modal LLMs on our tasks. InternVL denotes InternVL-Chat-V1-5 and LLaVA denotes LLaVA-NeXT-34B. Gemini FTE refers to the Gemini 2.0 Flash Thinking Experimental model. We use \emph{italic} fonts to refer to the recent thinking models.}
    \vspace{-0.2in}
    \label{tab:performance_matrix_format}
\end{table}

\paragraph{Results}
The top (text-only) section of Table~\ref{tab:performance_matrix_format} presents the results.
All the LLMs perform poorly on the three sets of our \datasetnamens. Notably, GPT-3.5, Mistral, and Llama-3 failed to show significant improvement over random performance.
Even for the remarkable GPT-4, GPT-4o and GPT-4v, their performance is not descent and particularly there is a huge performance gap between them and humans. 

In addition, as our \datasetname is essentially an inductive reasoning task from grid-represented examples, we also tested the thinking (o1-style) models concurrently with our work. We experimented with {\small \texttt{gemini-2.0-flash-thinking-exp-1219}}, {\small \texttt{o1-2024-12-17}}, {\small \texttt{o3-mini-2025-01-31}} and {\small \texttt{DeepSeek-R1}}. The former two models support multimodal inputs.
Because o1 is very slow and especially has a limited quota, we first compare it on a subset of 50 instances for both text and multimodal input. This preliminary study gives an accuracy of 42.0 (text) and 46.0 (visual), where GPT-4 (text) performs {44.0}.
We therefore run the full experiment with o1 (visual) only.
The reasoning models indeed achieve better results in the text-only setting, but fails to significantly improve over GPT-4o. 
The detailed performance decomposition of GPT-4, GPT-4o, o1 and Gemini FTE can be found in Appendix \ref{app:perf_decomp}.

We conducted an in-depth investigation into the discrepancy in the performance of DeepSeek-R1, which achieves strong results on \textsc{Assoc.}-Test but performs poorly on \textsc{Core}-Test. It revealed that R1 tends to develop transformation rules based on physical concepts and then applies these rules to predict the exact outputs from given inputs.
While this strategy works well for ARC examples due to their deterministic nature, the \textsc{Core} examples typically lack this property. We believe that tuning the prompts to provide better guidance could help mitigate this issue, which we leave for future work.

\paragraph{Conclusion}
Comparing the human performance in \ref{rq:human_perf} to the best-performing LLMs reveals a huge gap. 
While these tasks are simple or trivial for humans, LLMs face substantial challenges, indicating a lack of deep understanding.

When comparing LLMs' performance on low-level natural language tasks in \ref{rq:textual_input} to  high-level abstract pattern understanding tasks, we observe significant declines.
This highlights the presence of the \emph{stochastic parrot} phenomenon in LLMs. 
Our dataset also \emph{quantifies the severity of this phenomenon}.
For example, while GPT-3.5 performs on par with GPT-4 on the low-level tasks, it nearly drops to random guessing on our high-level tasks, revealing its tendency to act as a stochastic parrot with the physical concepts in our dataset.

\rqsection{Can multimodal LLMs perform well on our tasks with visual input representations?}
\label{rq:visual_input}

Next, we explore whether multi-modal LLMs can effectively solve our tasks when the input examples are presented as visual images rather than matrices like in \ref{rq:matrix_input}. We use the prompt in Figure~\ref{fig:visual_prompt_template} to query the answers from evaluated LLMs.

\paragraph{Results}
The bottom (multi-modal) section of Table~\ref{tab:performance_matrix_format} shows the results.
Consistent with the observations in \ref{rq:matrix_input}, a significant gap between the performance of LLMs and humans exists.

Notably, the recently introduced GPT-4o outperforms all other LLMs on \coredatasetns\ by 10\% with visual inputs but lags behind GPT-4 on matrix inputs. 
This discrepancy may be due to GPT-4o's training on images that directly illustrate physical concepts, making it more adept at solving problems like in Figure~\ref{fig:teaser}. However, this advantage does not extend to the more abstract problems in \harddatasetns\ that require further knowledge application skills, highlighting the LLMs' lack of deep understanding even with multi-modal training.

Finally, given that LLMs can generate high-quality descriptions of the concepts (see \ref{rq:textual_input}), we adopt a chain-of-thought~\cite{wei2022chain} approach.
It first asks the LLMs to describe each choice and then makes predictions.
The results in Table~\ref{tab:performance_matrix_format} (+CoT) show limited improvement or performance drop, further confirming the presence of the stochastic parrot phenomenon.

\rqsection{Is \datasetname challenging mainly due to LLMs'
unfamiliarity with grid representations?}
\label{rq:format_analysis}
{%
One might argue that the challenges of \datasetname might be due to the uncommon nature of the task format (especially the matrix-format inputs) encountered during LLM training, rather than a lack of deep understanding. 
We disprove this hypothesis from two perspectives:

\emph{(1) We show that \textbf{GPT-4o is actually {familiar with} the grid representations to some extent}.}
Specifically, we conducted a human study to examine GPT-4o's fundamental visual comprehension skills~\cite{girshick2014rich,long2015fully,he2017mask}, including recognizing objects from the grids, describing their colors and shapes, and identifying which objects have their color, shape, or position changed from input to output. 
These tasks correspond to the fundamental computer vision tasks of segmentation and object detection. 

We sampled 60 grid pairs from our dataset and had 3 annotators determine if GPT-4o provides correct answers. For each object, the answer is counted as correct only if the shape, color, and positions are all answered correctly. Our results show an accuracy of 86.7\%, which is significantly better compared to the accuracy on our high-level tasks. This confirms that GPT-4o is indeed familiar with the grid inputs, which aligns with the findings of the concurrent study~\cite{fluid}, but still cannot handle our \datasetname tasks effectively. 

Additionally, we studied Chain-of-Thoughts with the low-level understanding results as intermediate steps. 
On the \coredatasetns-Dev set,
the results below show that it still fails to improve over the vanilla GPT-4o prompting, showing that the GPT-4o  model cannot connect the low-level understanding with high-level concepts.

\begin{table}[h!]
    \small
    \centering
    \vspace{-0.1in}
    \resizebox{\columnwidth}{!}{
    \begin{tabular}{l c||lc}
    \toprule
    CoT - definitions &46.0$_{\pm\text{2.5}}$  &CoT - low-level &50.7$_{\pm\text{0.5}}$  \\
    \bottomrule
    \end{tabular}}
    \vspace{-0.1in}
    \vspace{-0.1in}
\end{table}

\emph{(2) We show that \textbf{making the LLMs more familiar with the grid representations does not lead to significant improvement}.} Specifically, we conduct the following experiments with text-only LLMs:
\begin{table}[h!]
    \small
    \centering
    \begin{tabular}{l cc}
    \toprule
     \bf Models   &  \bf \textsc{Core} & \bf \textsc{Assoc.}\\
    \midrule
    GPT-4 & 41.3$_{\pm\text{1.3}}$& 39.0$_{\pm\text{0.6}}$ \\
    \quad w/ ICL-3-shot &39.5$_{\pm\text{1.6}}$ &36.2$_{\pm\text{1.7}}$  \\
    \quad w/ ICL-9-shot &32.8$_{\pm\text{1.0}}$  & 39.0$_{\pm\text{1.6}}$ \\
    \midrule
    Mistral& 21.5$_{\pm\text{0.3}}$ & 23.2$_{\pm\text{0.4}}$\\
    \quad w/ FT on syn-tasks & 20.9$_{\pm\text{0.7}}$ &22.5$_{\pm\text{0.5}}$ \\
    \quad w/ FT on ARC & 20.9$_{\pm\text{0.8}}$& 25.5$_{\pm\text{0.9}}$\\
    \midrule    
    Llama-3 &  23.5$_{\pm\text{2.5}}$ & 21.7$_{\pm\text{2.0}}$ \\
    \quad w/ FT on syn-tasks & 23.0$_{\pm\text{1.1}}$ & 23.2$_{\pm\text{2.7}}$ \\
    \quad w/ FT on ARC & 22.2$_{\pm\text{1.6}}$ & 22.4$_{\pm\text{1.2}}$ \\
    \bottomrule
    \end{tabular}
    \caption{Performance of LLMs with in-context learning or fine-tuning on grid-format data.}
    \label{tab:performance_unsup_training}
    \vspace{-0.3in}
\end{table}

\begin{itemize}[noitemsep,nolistsep,leftmargin=*]
    \item \emph{ICL on other concepts. }
    Compare the performance of zero-shot GPT-4 with GPT-4 using in-context learning (ICL) on few-shot examples from concepts other than the assessed one.
    \item \emph{FT on synthetic matrix data. } Compare the open-source LLMs before and after fine-tuning on a large amount of matrix-input data (Appendix~\ref{app:synthetic_data})
    \item \emph{FT on the ARC task. } 
    Compare the open-source LLMs before and after fine-tuning on the original ARC~\cite{chollet2019measure} task, which ensures that all inputs from the \harddatasetns\ examples have been seen during model training.
\end{itemize}
Despite that both the ICL and SFT approaches make LLMs more familiar with matrix-format inputs, neither approach significantly improves the results as shown in Table~\ref{tab:performance_unsup_training}.

\paragraph{Conclusion} GPT-4o is somehow familiar with the grid format and further enhancing the familiarity of grid format for LLMs is not the key to addressing our challenges.

}

\rqsection{How much can LLMs benefit from training on labeled data?}
\label{rq:sup_training}

Many tasks that challenge LLMs can see significant performance boosts through ICL or SFT on labeled data~\cite{hessel-etal-2023-androids,yu2023personality,berglund2023reversal}. 
When such improvements are observed, it suggests that LLMs inherently possess the necessary skills to excel in their tasks, needing only minimal training effort.

We demonstrate that this is not the case for our tasks, where light-weight training of LLMs on labeled data does not improve for our tasks. Given the current lack of large-scale training data for our purpose, we conduct an extreme case study: models learn from the same concepts in \coredataset and are tested on the same concepts in \harddatasetns.
To this end, we select the instances that consists of at least two choices that exist in the \coredatasetns, leaving 80 examples.

\begin{table*}[t!]
    \small
    \centering
    \begin{tabular}{lc||lc||lc}
    \toprule
    GPT-4 & 42.9$_{\pm\text{2.4}}$ & GPT-4o & 40.4$_{\pm\text{2.1}}$ & Llama-3 & 22.1$_{\pm\text{2.8}}$ \\
    \,\,+ ICL on \textsc{Core}  &40.0$_{\pm\text{1.0}}$ & \,\,+ ICL on \textsc{Core}  &37.1$_{\pm\text{2.6}}$ &\,\,+ SFT on \textsc{Core} & 20.9$_{\pm\text{2.7}}$  \\
    \bottomrule
    \end{tabular}
    \caption{Accuracy on the subset of \textsc{Associative} subtask that has overlapped concepts with \textsc{Core}.}\label{tab:performance_sup_training}
    \vspace{-0.2in}
\end{table*}

We conduct the following experiments on this subset to answer \ref{rq:sup_training}:

\begin{itemize}[noitemsep,nolistsep,leftmargin=*]
    \item \emph{ICL on the same concepts. } Compare the zero-shot GPT-4/4o and GPT-4/4o with ICL\footnote{For GPT-4o, we implement ICL with multi-turn dialogues. Each shot in the demonstration is provided in one turn which asks the GPT-4o to explain the image.} on examples for the same concepts from \coredatasetns. Specifically, for each instance, we sample 9 examples from \coredatasetns\ with their labels among the choices of the instance.
    \item \emph{SFT on the \textsc{Core} set. } Compare the open-source LLMs before and after fine-tuning on labeled data from \coredatasetns.
\end{itemize}

\paragraph{Results} 

Table~\ref{tab:performance_sup_training} shows that ICL and SFT on the labeled examples of the same concepts lead to a consistent, though not severe, drop in performance.
The results suggest that the models have become overfitted to the "clean" examples from the \coredatasetns. They appear to have learned superficial correlations from the demonstrations that do not generalize well, providing further evidence of the stochastic parrot phenomenon.
The difficulty of generalization \emph{within the same concepts} indicates the challenges of our tasks to the supervised fine-tuning paradigm.

\paragraph{Conclusion} Together with the results for \ref{rq:format_analysis} and \ref{rq:sup_training}, it suggests that the low performance of LLMs is not likely to be improved from prompting techniques alone. There exists intrinsic inefficiency in the pre-training of LLMs, which results in the lack of necessary skills for deep understanding.

\section{Related Work}

\paragraph{Stochastic Parrots on LLMs}
The pioneer study by \cite{bender2020climbing} questioned the understanding ability of large models; and \citet{bender2021dangers} first introduced the terminology of stochastic parrot.
The concept of stochastic parrot has received great attention, leading to a surge of studies on this topic. 
According to Google Scholar, the term ``stochastic parrot'' appears in the titles of dozens of papers from diverse research fields~\cite{borji2023stochastic,li2023dark,duan2024flocks,henrique2023stochastic}. 
However, although the concept of stochastic parrots in LLMs is widely accepted and recognized, to the best of our knowledge, there is a lack of quantitative experiments to precisely verify this viewpoint. This gap directly motivates our work.

\paragraph{Abstract Reasoning Challenge}

Abstract reasoning challenge (ARC) aims to examine the inductive reasoning ability in a few-shot scenario~\cite{chollet2019measure} and it has been used as a remarkable testbed to measure the intelligence of LLMs.
Recently, many research efforts have been made on improving the performance of LLMs on ARC benchmark~\cite{tan2023large,wang2023hypothesis,xu2023llms,mirchandani2023large,wang2024speak,huang2024anpl}. 
We draw inspiration from ARC by utilizing input-output grids as abstract representations in our task design. However, our task is significantly different from the ARC-style work --- our high-level understanding task focuses on comprehending the transformation rules from inputs to outputs and relating them to physical concepts, and is designed to assess the stochastic parrot phenomenon.

\paragraph{Challenging Tasks towards LLMs' Understanding}
Extensive recent efforts have been made on designing tasks that challenge the understanding abilities of LLMs~\cite{chakrabarty-etal-2022-flute,tong2024metaphor,shapira-etal-2023-well,hessel-etal-2023-androids,donadel2024can,li2024previously}. 
For example, %
~\citet{hessel-etal-2023-androids} proposed a humor understanding task, revealing a large performance gap between LLMs and humans.
As a by-product, our \datasetname challenges the understanding capabilities of LLMs, relating it to the above studies. 
However, we make primary contribution to provide an quantitative experiment to verify stochastic parrots in LLMs via controllably paired low-level and high-level tasks.

\section{Conclusion}
We introduce \datasetnamens, a novel task to assess
machines' understanding of physical concepts at different levels.
Our experiments reveal that:
1) LLMs lag significantly behind humans on \datasetnamens, indicating a lack of deep understanding of the covered concepts;
2) LLMs exhibit the stochastic parrot phenomenon, as they excel at low-level remembering tasks but struggle with high-level understanding tasks;
3) LLMs' poor performance stems from its intrinsic deficiencies, as neither in-context learning nor fine-tuning improves their results.

\section*{Acknowledgment}
We thank the anonymous reviewers for their constructive feedback.
We also express our gratitude to Mr. François Chollet for developing the ARC benchmark and the annotation tool for abstract grid tasks\footnote{\url{https://arc-editor.lab42.global/?next=\%2Feditor}}. His introduction of this tool to us was particularly instrumental in the creation of \datasetnamens. This work has also been made possible by a Research Impact Fund project (RIF R6003-21) and a General Research Fund project (GRF 16203224) funded by the Research Grants Council (RGC) of the Hong Kong Government.

\bibliography{custom}

\begin{thebibliography}{53}
\providecommand{\natexlab}[1]{#1}

\bibitem[{Achiam et~al.(2023)Achiam, Adler, Agarwal, Ahmad, Akkaya, Aleman, Almeida, Altenschmidt, Altman, Anadkat et~al.}]{achiam2023gpt}
Josh Achiam, Steven Adler, Sandhini Agarwal, Lama Ahmad, Ilge Akkaya, Florencia~Leoni Aleman, Diogo Almeida, Janko Altenschmidt, Sam Altman, Shyamal Anadkat, et~al. 2023.
\newblock Gpt-4 technical report.
\newblock \emph{arXiv preprint arXiv:2303.08774}.

\bibitem[{Acquaviva et~al.(2022)Acquaviva, Pu, Kryven, Sechopoulos, Wong, Ecanow, Nye, Tessler, and Tenenbaum}]{acquaviva2022communicating}
Sam Acquaviva, Yewen Pu, Marta Kryven, Theodoros Sechopoulos, Catherine Wong, Gabrielle Ecanow, Maxwell Nye, Michael Tessler, and Josh Tenenbaum. 2022.
\newblock Communicating natural programs to humans and machines.
\newblock \emph{Advances in Neural Information Processing Systems}, 35:3731--3743.

\bibitem[{Armstrong(2010)}]{armstrong2010bloom}
Patricia Armstrong. 2010.
\newblock Bloom’s taxonomy.
\newblock \emph{Vanderbilt University Center for Teaching}, pages 1--3.

\bibitem[{Banerjee and Lavie(2005)}]{banerjee2005meteor}
Satanjeev Banerjee and Alon Lavie. 2005.
\newblock Meteor: An automatic metric for mt evaluation with improved correlation with human judgments.
\newblock In \emph{Proceedings of the acl workshop on intrinsic and extrinsic evaluation measures for machine translation and/or summarization}, pages 65--72.

\bibitem[{Bender et~al.(2021)Bender, Gebru, McMillan-Major, and Shmitchell}]{bender2021dangers}
Emily~M Bender, Timnit Gebru, Angelina McMillan-Major, and Shmargaret Shmitchell. 2021.
\newblock On the dangers of stochastic parrots: Can language models be too big?
\newblock In \emph{Proceedings of the 2021 ACM conference on fairness, accountability, and transparency}, pages 610--623.

\bibitem[{Bender and Koller(2020)}]{bender2020climbing}
Emily~M Bender and Alexander Koller. 2020.
\newblock Climbing towards nlu: On meaning, form, and understanding in the age of data.
\newblock In \emph{Proceedings of the 58th annual meeting of the association for computational linguistics}, pages 5185--5198.

\bibitem[{Berglund et~al.(2023)Berglund, Tong, Kaufmann, Balesni, Stickland, Korbak, and Evans}]{berglund2023reversal}
Lukas Berglund, Meg Tong, Max Kaufmann, Mikita Balesni, Asa~Cooper Stickland, Tomasz Korbak, and Owain Evans. 2023.
\newblock The reversal curse: Llms trained on" a is b" fail to learn" b is a".
\newblock \emph{arXiv preprint arXiv:2309.12288}.

\bibitem[{Black and Wiliam(1998{\natexlab{a}})}]{black1998assessment}
Paul Black and Dylan Wiliam. 1998{\natexlab{a}}.
\newblock Assessment and classroom learning.
\newblock \emph{Assessment in Education: principles, policy \& practice}, 5(1):7--74.

\bibitem[{Black and Wiliam(1998{\natexlab{b}})}]{black1998inside}
Paul Black and Dylan Wiliam. 1998{\natexlab{b}}.
\newblock \emph{Inside the black box: Raising standards through classroom assessment}.
\newblock Granada Learning.

\bibitem[{Bommasani et~al.(2021)Bommasani, Hudson, Adeli, Altman, Arora, von Arx, Bernstein, Bohg, Bosselut, Brunskill et~al.}]{bommasani2021opportunities}
Rishi Bommasani, Drew~A Hudson, Ehsan Adeli, Russ Altman, Simran Arora, Sydney von Arx, Michael~S Bernstein, Jeannette Bohg, Antoine Bosselut, Emma Brunskill, et~al. 2021.
\newblock On the opportunities and risks of foundation models.
\newblock \emph{arXiv preprint arXiv:2108.07258}.

\bibitem[{Borji(2023)}]{borji2023stochastic}
Ali Borji. 2023.
\newblock Stochastic parrots or intelligent systems? a perspective on true depth of understanding in llms.
\newblock \emph{A Perspective on True Depth of Understanding in LLMs (July 11, 2023)}.

\bibitem[{Brown et~al.(2020)Brown, Mann, Ryder, Subbiah, Kaplan, Dhariwal, Neelakantan, Shyam, Sastry, Askell et~al.}]{brown2020language}
Tom Brown, Benjamin Mann, Nick Ryder, Melanie Subbiah, Jared~D Kaplan, Prafulla Dhariwal, Arvind Neelakantan, Pranav Shyam, Girish Sastry, Amanda Askell, et~al. 2020.
\newblock Language models are few-shot learners.
\newblock \emph{Advances in neural information processing systems}, 33:1877--1901.

\bibitem[{Chakrabarty et~al.(2022)Chakrabarty, Saakyan, Ghosh, and Muresan}]{chakrabarty-etal-2022-flute}
Tuhin Chakrabarty, Arkadiy Saakyan, Debanjan Ghosh, and Smaranda Muresan. 2022.
\newblock \href {https://doi.org/10.18653/v1/2022.emnlp-main.481} {{FLUTE}: Figurative language understanding through textual explanations}.
\newblock In \emph{Proceedings of the 2022 Conference on Empirical Methods in Natural Language Processing}, pages 7139--7159, Abu Dhabi, United Arab Emirates. Association for Computational Linguistics.

\bibitem[{Chang et~al.(2024)Chang, Wang, Wang, Wu, Yang, Zhu, Chen, Yi, Wang, Wang et~al.}]{chang2024survey}
Yupeng Chang, Xu~Wang, Jindong Wang, Yuan Wu, Linyi Yang, Kaijie Zhu, Hao Chen, Xiaoyuan Yi, Cunxiang Wang, Yidong Wang, et~al. 2024.
\newblock A survey on evaluation of large language models.
\newblock \emph{ACM Transactions on Intelligent Systems and Technology}, 15(3):1--45.

\bibitem[{Chen et~al.(2024)Chen, Wang, Tian, Ye, Gao, Cui, Tong, Hu, Luo, Ma et~al.}]{chen2024far}
Zhe Chen, Weiyun Wang, Hao Tian, Shenglong Ye, Zhangwei Gao, Erfei Cui, Wenwen Tong, Kongzhi Hu, Jiapeng Luo, Zheng Ma, et~al. 2024.
\newblock How far are we to gpt-4v? closing the gap to commercial multimodal models with open-source suites.
\newblock \emph{arXiv preprint arXiv:2404.16821}.

\bibitem[{Chen et~al.(2023)Chen, Wu, Wang, Su, Chen, Xing, Zhong, Zhang, Zhu, Lu, Li, Luo, Lu, Qiao, and Dai}]{chen2023internvl}
Zhe Chen, Jiannan Wu, Wenhai Wang, Weijie Su, Guo Chen, Sen Xing, Muyan Zhong, Qinglong Zhang, Xizhou Zhu, Lewei Lu, Bin Li, Ping Luo, Tong Lu, Yu~Qiao, and Jifeng Dai. 2023.
\newblock Internvl: Scaling up vision foundation models and aligning for generic visual-linguistic tasks.
\newblock \emph{arXiv preprint arXiv:2312.14238}.

\bibitem[{Chollet(2019)}]{chollet2019measure}
Fran{\c{c}}ois Chollet. 2019.
\newblock On the measure of intelligence.
\newblock \emph{arXiv preprint arXiv:1911.01547}.

\bibitem[{Demszky et~al.(2023)Demszky, Yang, Yeager, Bryan, Clapper, Chandhok, Eichstaedt, Hecht, Jamieson, Johnson et~al.}]{demszky2023using}
Dorottya Demszky, Diyi Yang, David~S Yeager, Christopher~J Bryan, Margarett Clapper, Susannah Chandhok, Johannes~C Eichstaedt, Cameron Hecht, Jeremy Jamieson, Meghann Johnson, et~al. 2023.
\newblock Using large language models in psychology.
\newblock \emph{Nature Reviews Psychology}, 2(11):688--701.

\bibitem[{Donadel et~al.(2024)Donadel, Marchiori, Pajola, and Conti}]{donadel2024can}
Denis Donadel, Francesco Marchiori, Luca Pajola, and Mauro Conti. 2024.
\newblock Can llms understand computer networks? towards a virtual system administrator.
\newblock \emph{arXiv preprint arXiv:2404.12689}.

\bibitem[{Duan et~al.(2024)Duan, Dziedzic, Papernot, and Boenisch}]{duan2024flocks}
Haonan Duan, Adam Dziedzic, Nicolas Papernot, and Franziska Boenisch. 2024.
\newblock Flocks of stochastic parrots: Differentially private prompt learning for large language models.
\newblock \emph{Advances in Neural Information Processing Systems}, 36.

\bibitem[{Duan et~al.(2022)Duan, Yu, Tan, Zhu, and Tan}]{duan2022survey}
Jiafei Duan, Samson Yu, Hui~Li Tan, Hongyuan Zhu, and Cheston Tan. 2022.
\newblock A survey of embodied ai: From simulators to research tasks.
\newblock \emph{IEEE Transactions on Emerging Topics in Computational Intelligence}, 6(2):230--244.

\bibitem[{Girshick et~al.(2014)Girshick, Donahue, Darrell, and Malik}]{girshick2014rich}
Ross Girshick, Jeff Donahue, Trevor Darrell, and Jitendra Malik. 2014.
\newblock Rich feature hierarchies for accurate object detection and semantic segmentation.
\newblock In \emph{Proceedings of the IEEE conference on computer vision and pattern recognition}, pages 580--587.

\bibitem[{Harlen and James(1997)}]{harlen1997assessment}
Wynne Harlen and Mary James. 1997.
\newblock Assessment and learning: differences and relationships between formative and summative assessment.
\newblock \emph{Assessment in education: Principles, policy \& practice}, 4(3):365--379.

\bibitem[{He et~al.(2017)He, Gkioxari, Doll{\'a}r, and Girshick}]{he2017mask}
Kaiming He, Georgia Gkioxari, Piotr Doll{\'a}r, and Ross Girshick. 2017.
\newblock Mask r-cnn.
\newblock In \emph{Proceedings of the IEEE international conference on computer vision}, pages 2961--2969.

\bibitem[{Henrique et~al.(2023)Henrique, Kucharavy, and Guerraoui}]{henrique2023stochastic}
Da~Silva~Gameiro Henrique, Andrei Kucharavy, and Rachid Guerraoui. 2023.
\newblock Stochastic parrots looking for stochastic parrots: Llms are easy to fine-tune and hard to detect with other llms.
\newblock \emph{arXiv preprint arXiv:2304.08968}.

\bibitem[{Hessel et~al.(2023)Hessel, Marasovic, Hwang, Lee, Da, Zellers, Mankoff, and Choi}]{hessel-etal-2023-androids}
Jack Hessel, Ana Marasovic, Jena~D. Hwang, Lillian Lee, Jeff Da, Rowan Zellers, Robert Mankoff, and Yejin Choi. 2023.
\newblock \href {https://doi.org/10.18653/v1/2023.acl-long.41} {Do androids laugh at electric sheep? humor {``}understanding{''} benchmarks from the new yorker caption contest}.
\newblock In \emph{Proceedings of the 61st Annual Meeting of the Association for Computational Linguistics (Volume 1: Long Papers)}, pages 688--714, Toronto, Canada. Association for Computational Linguistics.

\bibitem[{Hu et~al.(2021)Hu, Wallis, Allen-Zhu, Li, Wang, Wang, Chen et~al.}]{hu2021lora}
Edward~J Hu, Phillip Wallis, Zeyuan Allen-Zhu, Yuanzhi Li, Shean Wang, Lu~Wang, Weizhu Chen, et~al. 2021.
\newblock Lora: Low-rank adaptation of large language models.
\newblock In \emph{International Conference on Learning Representations}.

\bibitem[{Huang et~al.(2024)Huang, Nan, Hu, Jin, Peng, Wen, Zhang, Du, Guo, Pu et~al.}]{huang2024anpl}
Di~Huang, Ziyuan Nan, Xing Hu, Pengwei Jin, Shaohui Peng, Yuanbo Wen, Rui Zhang, Zidong Du, Qi~Guo, Yewen Pu, et~al. 2024.
\newblock Anpl: Towards natural programming with interactive decomposition.
\newblock \emph{Advances in Neural Information Processing Systems}, 36.

\bibitem[{Jiang et~al.(2023)Jiang, Sablayrolles, Mensch, Bamford, Chaplot, Casas, Bressand, Lengyel, Lample, Saulnier et~al.}]{jiang2023mistral}
Albert~Q Jiang, Alexandre Sablayrolles, Arthur Mensch, Chris Bamford, Devendra~Singh Chaplot, Diego de~las Casas, Florian Bressand, Gianna Lengyel, Guillaume Lample, Lucile Saulnier, et~al. 2023.
\newblock Mistral 7b.
\newblock \emph{arXiv preprint arXiv:2310.06825}.

\bibitem[{Krathwohl(2002)}]{krathwohl2002revision}
David~R Krathwohl. 2002.
\newblock A revision of bloom's taxonomy: An overview.
\newblock \emph{Theory into practice}, 41(4):212--218.

\bibitem[{Li et~al.(2024)Li, Wang, Xu, Pang, Yu, Lin, Wang, and Zhou}]{li2024previously}
Jiangnan Li, Qiujing Wang, Liyan Xu, Wenjie Pang, Mo~Yu, Zheng Lin, Weiping Wang, and Jie Zhou. 2024.
\newblock Previously on the stories: Recap snippet identification for story reading.
\newblock \emph{arXiv preprint arXiv:2402.07271}.

\bibitem[{Li(2023)}]{li2023dark}
Zihao Li. 2023.
\newblock The dark side of chatgpt: legal and ethical challenges from stochastic parrots and hallucination.
\newblock \emph{arXiv preprint arXiv:2304.14347}.

\bibitem[{Liu et~al.(2023{\natexlab{a}})Liu, Li, Li, and Lee}]{liu2023improvedllava}
Haotian Liu, Chunyuan Li, Yuheng Li, and Yong~Jae Lee. 2023{\natexlab{a}}.
\newblock Improved baselines with visual instruction tuning.

\bibitem[{Liu et~al.(2023{\natexlab{b}})Liu, Li, Wu, and Lee}]{liu2023llava}
Haotian Liu, Chunyuan Li, Qingyang Wu, and Yong~Jae Lee. 2023{\natexlab{b}}.
\newblock Visual instruction tuning.
\newblock In \emph{NeurIPS}.

\bibitem[{Long et~al.(2015)Long, Shelhamer, and Darrell}]{long2015fully}
Jonathan Long, Evan Shelhamer, and Trevor Darrell. 2015.
\newblock Fully convolutional networks for semantic segmentation.
\newblock In \emph{Proceedings of the IEEE conference on computer vision and pattern recognition}, pages 3431--3440.

\bibitem[{MetaAI(2024)}]{llama3}
MetaAI. 2024.
\newblock \href {https://ai.meta.com/blog/meta-llama-3/} {Introducing meta llama 3: The most capable openly available llm to date}.

\bibitem[{Min et~al.(2023)Min, Ross, Sulem, Veyseh, Nguyen, Sainz, Agirre, Heintz, and Roth}]{min2023recent}
Bonan Min, Hayley Ross, Elior Sulem, Amir Pouran~Ben Veyseh, Thien~Huu Nguyen, Oscar Sainz, Eneko Agirre, Ilana Heintz, and Dan Roth. 2023.
\newblock Recent advances in natural language processing via large pre-trained language models: A survey.
\newblock \emph{ACM Computing Surveys}, 56(2):1--40.

\bibitem[{Mirchandani et~al.(2023)Mirchandani, Xia, Florence, Ichter, Driess, Arenas, Rao, Sadigh, and Zeng}]{mirchandani2023large}
Suvir Mirchandani, Fei Xia, Pete Florence, Brian Ichter, Danny Driess, Montserrat~Gonzalez Arenas, Kanishka Rao, Dorsa Sadigh, and Andy Zeng. 2023.
\newblock \href {https://arxiv.org/abs/2307.04721} {Large language models as general pattern machines}.
\newblock \emph{ArXiv preprint}, abs/2307.04721.

\bibitem[{Mitchell and Krakauer(2023)}]{mitchell2023debate}
Melanie Mitchell and David~C Krakauer. 2023.
\newblock The debate over understanding in ai’s large language models.
\newblock \emph{Proceedings of the National Academy of Sciences}, 120(13):e2215907120.

\bibitem[{Papineni et~al.(2002)Papineni, Roukos, Ward, and Zhu}]{papineni2002bleu}
Kishore Papineni, Salim Roukos, Todd Ward, and Wei-Jing Zhu. 2002.
\newblock Bleu: a method for automatic evaluation of machine translation.
\newblock In \emph{Proceedings of the 40th annual meeting of the Association for Computational Linguistics}, pages 311--318.

\bibitem[{Peng et~al.(2024)Peng, Zheng, Liu, Han, Wang, Yang, Song, Zhao, Wei, Li et~al.}]{peng2024large}
Di~Peng, Liubin Zheng, Dan Liu, Cheng Han, Xin Wang, Yan Yang, Li~Song, Miaoying Zhao, Yanfeng Wei, Jiayi Li, et~al. 2024.
\newblock Large-language models facilitate discovery of the molecular signatures regulating sleep and activity.
\newblock \emph{Nature Communications}, 15(1):3685.

\bibitem[{Savva et~al.(2019)Savva, Kadian, Maksymets, Zhao, Wijmans, Jain, Straub, Liu, Koltun, Malik et~al.}]{savva2019habitat}
Manolis Savva, Abhishek Kadian, Oleksandr Maksymets, Yili Zhao, Erik Wijmans, Bhavana Jain, Julian Straub, Jia Liu, Vladlen Koltun, Jitendra Malik, et~al. 2019.
\newblock Habitat: A platform for embodied ai research.
\newblock In \emph{Proceedings of the IEEE/CVF international conference on computer vision}, pages 9339--9347.

\bibitem[{Shapira et~al.(2023)Shapira, Zwirn, and Goldberg}]{shapira-etal-2023-well}
Natalie Shapira, Guy Zwirn, and Yoav Goldberg. 2023.
\newblock \href {https://doi.org/10.18653/v1/2023.findings-acl.663} {How well do large language models perform on faux pas tests?}
\newblock In \emph{Findings of the Association for Computational Linguistics: ACL 2023}, pages 10438--10451, Toronto, Canada. Association for Computational Linguistics.

\bibitem[{Tan and Motani(2023)}]{tan2023large}
John Chong~Min Tan and Mehul Motani. 2023.
\newblock \href {https://arxiv.org/abs/2310.05146} {Large language model (llm) as a system of multiple expert agents: An approach to solve the abstraction and reasoning corpus (arc) challenge}.
\newblock \emph{ArXiv preprint}, abs/2310.05146.

\bibitem[{Team et~al.(2023)Team, Anil, Borgeaud, Wu, Alayrac, Yu, Soricut, Schalkwyk, Dai, Hauth et~al.}]{team2023gemini}
Gemini Team, Rohan Anil, Sebastian Borgeaud, Yonghui Wu, Jean-Baptiste Alayrac, Jiahui Yu, Radu Soricut, Johan Schalkwyk, Andrew~M Dai, Anja Hauth, et~al. 2023.
\newblock Gemini: a family of highly capable multimodal models.
\newblock \emph{arXiv preprint arXiv:2312.11805}.

\bibitem[{Tong et~al.(2024)Tong, Choenni, Lewis, and Shutova}]{tong2024metaphor}
Xiaoyu Tong, Rochelle Choenni, Martha Lewis, and Ekaterina Shutova. 2024.
\newblock Metaphor understanding challenge dataset for llms.
\newblock \emph{arXiv preprint arXiv:2403.11810}.

\bibitem[{Wang et~al.(2023)Wang, Zelikman, Poesia, Pu, Haber, and Goodman}]{wang2023hypothesis}
Ruocheng Wang, Eric Zelikman, Gabriel Poesia, Yewen Pu, Nick Haber, and Noah~D Goodman. 2023.
\newblock \href {https://arxiv.org/abs/2309.05660} {Hypothesis search: Inductive reasoning with language models}.
\newblock \emph{ArXiv preprint}, abs/2309.05660.

\bibitem[{Wang et~al.(2024)Wang, Cheng, Sun, Li, and Liu}]{wang2024speak}
Yile Wang, Sijie Cheng, Zixin Sun, Peng Li, and Yang Liu. 2024.
\newblock \href {https://arxiv.org/abs/2401.11725} {Speak it out: Solving symbol-related problems with symbol-to-language conversion for language models}.
\newblock \emph{ArXiv preprint}, abs/2401.11725.

\bibitem[{Wei et~al.(2022)Wei, Wang, Schuurmans, Bosma, Xia, Chi, Le, Zhou et~al.}]{wei2022chain}
Jason Wei, Xuezhi Wang, Dale Schuurmans, Maarten Bosma, Fei Xia, Ed~Chi, Quoc~V Le, Denny Zhou, et~al. 2022.
\newblock Chain-of-thought prompting elicits reasoning in large language models.
\newblock \emph{Advances in neural information processing systems}, 35:24824--24837.

\bibitem[{Wu et~al.(2025)Wu, Yu, Liu, Yeung, and Zhou}]{fluid}
Junjie Wu, Mo~Yu, Lemao Liu, Dit-Yan Yeung, and Jie Zhou. 2025.
\newblock \href {https://wujunjie1998.github.io/araoc-benchmark.github.io/} {Understanding llms’ fluid intelligence deficiency: An analysis of the arc task}.
\newblock In \emph{Proceedings of the 2025 Conference of the Nations of the Americas Chapter of the Association for Computational Linguistics}.

\bibitem[{Xiang et~al.(2023)Xiang, Tao, Gu, Shu, Wang, Yang, and Hu}]{xiang2024language}
Jiannan Xiang, Tianhua Tao, Yi~Gu, Tianmin Shu, Zirui Wang, Zichao Yang, and Zhiting Hu. 2023.
\newblock Language models meet world models: Embodied experiences enhance language models.
\newblock \emph{Advances in neural information processing systems}, 36.

\bibitem[{Xu et~al.(2023)Xu, Li, Vaezipoor, Sanner, and Khalil}]{xu2023llms}
Yudong Xu, Wenhao Li, Pashootan Vaezipoor, Scott Sanner, and Elias~B Khalil. 2023.
\newblock \href {https://arxiv.org/abs/2305.18354} {Llms and the abstraction and reasoning corpus: Successes, failures, and the importance of object-based representations}.
\newblock \emph{ArXiv preprint}, abs/2305.18354.

\bibitem[{Yu et~al.(2023)Yu, Li, Yao, Pang, Zhou, Xiao, Meng, and Zhou}]{yu2023personality}
Mo~Yu, Jiangnan Li, Shunyu Yao, Wenjie Pang, Xiaochen Zhou, Zhou Xiao, Fandong Meng, and Jie Zhou. 2023.
\newblock Personality understanding of fictional characters during book reading.
\newblock In \emph{Proceedings of the 61st Annual Meeting of the Association for Computational Linguistics (Volume 1: Long Papers)}, pages 14784--14802.

\end{thebibliography}

\appendix
\clearpage

\section{Details of the Included Concepts in our \datasetnamens}
\label{app:dataset_details}

\paragraph{Concepts in \coredatasetns}
The concepts in \coredatasetns\ are basic physical concepts that we manually design problems for. The development set covers 27 concepts and the test set covers 25 concepts as follows:

\begin{table}[h!]
    \small
    \centering
    \begin{adjustbox}{width=\linewidth}
    \begin{tabular}{l c||lc}
    \toprule
    reference frame&12&gravity&10\\
    reflection&10&refraction&10\\
    light imaging&10&communicating vessels&10\\
    cut&10&laser&10\\
    surface tension&10&move&10\\
    \midrule
    buoyancy&10&acceleration&10\\
    inertia&10&electricity&10\\
    repulsive force&8&wave&8\\
    lever&6&optical filters&6\\
    compression&4&diffuse reflection of light&4\\
    \midrule
    wave interference&4&diffusion&4\\
    vortex&4&expansion&4\\
    nuclear fission&2&nuclear fusion&2\\
    diffraction of waves&2\\
    \bottomrule
    \end{tabular}
    \end{adjustbox}
    \caption{Concepts and their corresponding number of instances in \coredatasetns-Dev.}
    \label{tab:concept_stats_core}
\end{table}

\begin{table}[h!]
    \small
    \centering
    \begin{adjustbox}{width=\linewidth}
    \begin{tabular}{l c||lc}
    \toprule
    atmospheric pressure & 12 & energe conservation & 10 \\
    elastic force & 10 & friction & 9 \\
    photoelectric effect & 8 & heat conduction & 8 \\
    doppler effect & 8 & electromagnetic wave & 8 \\
    melting & 8 & vaporization & 8 \\
    \midrule
    fluid pressure & 8 & thermal expansion and contraction & 8 \\
    Brownian motion & 8 & splashing & 8 \\
    oscillation & 8 & relativity & 8 \\
    lighting & 8 & lifting & 8 \\
    force composition & 8 & pulley & 8 \\
    \midrule
    inclined plane & 8 & Bernoulli effect & 7 \\
    fictitious force & 6 & siphon & 6 \\
    resonance & 4 & ~ & ~ \\
    \bottomrule
    \end{tabular}
    \end{adjustbox}
    \caption{Concepts and their corresponding number of instances in \coredatasetns-Test.}
    \label{tab:concept_stats_core_test}
\end{table}

\paragraph{Concepts in \harddatasetns}
The following table summarized all the concepts from \harddatasetns:

\begin{table}[h!]
    \small
    \centering
    \begin{adjustbox}{width=\linewidth}
    \begin{tabular}{l c||lc}
    \toprule
    mirror                    & 30 & laser                     & 20 \\
    zoom in                   & 15 & magnet                    & 14 \\
    wave                      & 13 & explosion                 & 11 \\
    compression               & 10 & rotation                  & 10 \\
    gravity                   &  9 & expansion                 &  9 \\
    \midrule
    move                      &  8 & change of reference frame &  8 \\
    water ripples             &  7 & long exposure             &  7 \\
    reflection                &  5 & wetting                   &  5 \\
    diffusion                 &  4 & zoom out                  &  3 \\
    projection                &  2 & polarization of light     &  1 \\
    \midrule
    vortex                    &  1 & chemical bond             &  1 \\
    nuclear fission           &  1 & squeeze                   &  1 \\
    nuclear fusion            &  1 & lumination                &  1 \\
    wave interference         &  1 & optical filter            &  1 \\
    vacuum                    &  1 &  \\
    \bottomrule
    \end{tabular}
    \end{adjustbox}
    \caption{Concepts and their corresponding number of instances in \harddatasetns.}
    \label{tab:concept_stats_all}
\end{table}

\section{Details of Analysis Methods in \ref{rq:textual_input}}
\label{app:rq1_details}
\subsection{Masking of Textual Descriptions}
This experiment follows the setting in the ``Physical Concept Selection Subtask'' in section \ref{sec: low-level}. The definitions of the corresponding phenomena were extracted from Wikipedia as well as generated by GPT-3.5 and GPT-4. To maintain consistency, the terms representing concepts were masked as {\small \texttt{[PHENOMENON]}} while relevant terms are masked as {\small \texttt{[MASK]}}. For instance, ``interference'' which corresponds to the phenomenon ``wave interference'' was masked as {\small \texttt{[PHENOMENON]}}. In contrast, ``Newton's first law of motion'' which corresponds to the phenomenon ``inertia'' was masked as {\small \texttt{[MASK]}}.

An example of the masked description can be found in Figure~\ref{fig:masked_description_example}.

\subsection{Prompts Used for Description Generation and Classification}
Figure~\ref{fig:nl_gen_prompt_template} and \ref{fig:nl_guess_prompt_template} include the prompts used for generation and classification respectively.

\begin{figure}[h]
    \centering
\begin{lstlisting}[language=prompt]
<<<[SYSTEM]>>>
You are an expert in physics. You task is to provide a comprehensive definition of a given physical concept or phenomenon, with the key properties or key examples of the concept included.

<<<[USER]>>>
Please provide me with the definition of the physical concept "{{ [[[CONCEPT]]] }}", with the key properties or key examples included.
\end{lstlisting}
    \caption{The prompt template used for generating descriptions of physical concepts (denoted as the variable \textcolor{magenta}{\small \textbf{\texttt{CONCEPT}}}) in \ref{rq:textual_input}.}
    \label{fig:nl_gen_prompt_template}
\end{figure}

\begin{figure}[h]
    \centering
\begin{lstlisting}[language=prompt]
<<<[SYSTEM]>>>
You will be playing a game: 
You are given a definition of a physical phenonmenon, where the names of the phenonmenon are masked.
Your task is to guess which phenonmenon the definiton refers to.
Please select the most close answer from the provided options.

<<<[USER]>>>
Here is a definition of a physical phenonmenon, where the names of the phenonmenon are masked:

===[Definition]===

{{ [[[MASKED DESCRIPTION]]] }}

Please guess which phenonmenon the definiton refers to. You should choose your answer from the following options: {{ [[[CANDIDATE ANSWERS]]] }}

Your response should end with your choice of answer.
\end{lstlisting}
    \caption{The prompt template used for guessing the referred physical concept from four candidates (denoted as the variable \textcolor{magenta}{\small \textbf{\texttt{CANDIDATE ANSWERS}}}) from the natural language descriptions (denoted as the variable \textcolor{magenta}{\small \textbf{\texttt{MASKED DESCRIPTION}}}) in \ref{rq:textual_input}.}
    \label{fig:nl_guess_prompt_template}
\end{figure}

\begin{figure*}[h]
    \centering
\begin{lstlisting}[language=prompt]
[PHENOMENON] is a fundamental concept in physics that describes the resistance of any physical object to a change in its state of motion. This concept is a central part of [MASK], often referred to as the law of [PHENOMENON]. According to this law, an object at rest will stay at rest, and an object in motion will continue to move at a constant velocity, unless acted upon by a net external force. Here are the key properties and examples of [PHENOMENON]:

### Key Properties:
1. **Dependence on Mass**: The [PHENOMENON] of an object is directly proportional to its mass. The greater the mass of an object, the greater its [PHENOMENON], and hence, the more force it requires to change its state of motion.

2. **Resistance to Acceleration**: [PHENOMENON] is essentially the resistance of an object to any change in its velocity, which includes changes in the speed or direction of the object's motion.

3. **Universal Applicability**: [PHENOMENON] applies to all objects with mass, whether they are microscopic or astronomical in scale.

4. **Independence from External Factors**: The [PHENOMENON] of an object is inherent and does not depend on external conditions such as the environment, temperature, or pressure.

### Key Examples:
1. **A Parked Car**: A parked car will not move unless a force is applied to it. Once moving, it will continue to move at a constant speed in a straight line unless forces like friction or brakes are applied to change its state.

2. **Astronauts and Objects in Space**: In the vacuum of space, where there is little to no external force, an astronaut or any other object will continue moving in the same direction and at the same speed until acted upon by another force. This is an example of [PHENOMENON] in a microgravity environment.

3. **Seatbelts in Vehicles**: When a car suddenly stops, the passengers inside tend to lurch forward. This is due to the [PHENOMENON] of their bodies; their bodies were in motion and tend to remain in motion despite the car stopping. Seatbelts provide the necessary force to counteract this [PHENOMENON] and keep the passengers safe.

4. **Tablecloth Trick**: A classic example demonstrating [PHENOMENON] is the tablecloth trick, where a quick pull of the tablecloth can leave dishes undisturbed on a table. The [PHENOMENON] of the dishes (their tendency to resist changes in motion) allows them to remain relatively still while the tablecloth is quickly pulled from under them.

Understanding [PHENOMENON] is crucial for analyzing the motion of objects in various physical contexts, from everyday life to complex scientific scenarios. It is a cornerstone in the study of dynamics and plays a critical role in engineering, automotive safety, aerospace technology, and many other fields.
\end{lstlisting}
    \caption{An example of our masked description for the concept \texttt{inertia}.}
    \label{fig:masked_description_example}
\end{figure*}

\subsection{Additional Results on the Self-Play Game}
\label{app:self_play}
Automatic evaluation of a text generation task is in general difficult.
Especially, in our scenario each concept have many different ground-truth examples in its description, thus existing automatic metrics such as BLEU~\cite{papineni2002bleu} and METEOR~\cite{banerjee2005meteor} are not capable of accurately measuring the quality. 
Therefore, we propose an alternative automatic metric via a self-play game for this subtask: 

For each generated description of a concept, we mask the synonyms of the concept in it as in the previous selection subtask, and ask the same LLM to identify the concept being described from four options. 
This metric evaluates the quality of LLMs' generated concept descriptions in an objective manner. 

\begin{table}[t!]
    \small
    \centering
    \vspace{-0.1in}
    \begin{tabular}{ccccc}
    \toprule
      & \bf Mistral & \bf Llama-3 & \bf GPT-3.5 & \bf GPT-4 \\
     \midrule
    Human &92.6& 100 &100 & 100\\
    \midrule
    SP & 89.2$_{\pm\text{1.6}}$ & 91.9$_{\pm\text{0.6}}$ &96.0$_{\pm\text{0.4}}$ & 99.8$_{\pm\text{0.2}}$\\    
    \bottomrule
    \end{tabular}
    \vspace{-0.1in}
    \caption{Evaluations on the concept generation subtask, with metrics of Self-Play success and Human evaluation.}
    \vspace{-0.1in}
    \label{tab:generation_extended}
\end{table}

\paragraph{Results}
The results of automatic evaluation via self-play are shown in Table~\ref{tab:generation_extended} together with the human evaluation results. 
In the self-play test, all LLMs can accurately recognize the physical concepts from the descriptions they wrote by themselves.
Combined with the conclusion from human evaluation, 
it shows the LLMs can generate correct and sufficient information.

\section{Details of the Methods Used in \ref{rq:matrix_input} and \ref{rq:visual_input}}

We use the prompt template in Figure~\ref{fig:matrix_prompt_template} for experiments on text-only LLMs (\ref{rq:matrix_input}); and the template in Figure~\ref{fig:visual_prompt_template} for multi-modal LLMs (\ref{rq:visual_input}).

\begin{figure*}[t]
    \centering
\begin{lstlisting}[language=prompt]
<<<[SYSTEM]>>>
You will be playing a game: 
You are given several examples. Each example consists of an``input grid'' and an ``output grid'' of numbers from 0-9, where each number corresponds to a color.
Your task if try to find the common patterns from the examples and abstract the meanings of the patterns in the physical or mathematics world.
Based on the recognized meaning, please select the most close description of the common pattern from the provided options.

<<<[USER]>>>
Lets play a game where you are transforming an input grid of numbers into an output grid of numbers.

The numbers represent different colors:
0 = black
1 = blue
2 = red
3 = green
4 = yellow
5 = gray
6 = magenta
7 = orange
8 = cyan
9 = brown

Here are examples of input grids and its corresponding output grids:

Example input grid:  
{{ [[[INPUT GRID1]]] }}

Example output grid:
{{ [[[OUTPUT GRID1]]] }}

Example input grid:  
{{ [[[INPUT GRID2]]] }}

Example output grid:
{{ [[[OUTPUT GRID2]]] }}

Example input grid:  
{{ [[[INPUT GRID3]]] }}

Example output grid:
{{ [[[OUTPUT GRID3]]] }}

Please first try to find the common patterns from the input-output pairs, then answer the following question:

What meanings in the physical or mathematics world can be abstracted from the patterns? Please choose your answer from the following options: 
{{ [[[CANDIDATE ANSWERS]]] }}

Your response should end with your choice of answer.
\end{lstlisting}
    \caption{The prompt template used in \ref{rq:matrix_input}. The pair of an \textcolor{magenta}{\small \textbf{\texttt{INPUT GRID}}} and an \textcolor{magenta}{\small \textbf{\texttt{OUTPUT GRID}}} consists of one example of a physical phenomenon in matrix format.}
    \label{fig:matrix_prompt_template}
\end{figure*}

\begin{figure*}[t]
    \centering
\begin{lstlisting}[language=prompt]
{{ [[[UPLOADED IMAGE]]] }}
<<<[USER]>>>
In the given image, there are two columns of matrices with elements represented by different colors. 
The left column represents the inputs, and the right column represents the corresponding outputs. 
For each row in the image, the output is derived from the input using the same transformation rule, 
which corresponds to a real-world physical concept.

Your task is to identify the physical concept demonstrated in this image from the following options:

{{ [[[CANDIDATE ANSWERS]]] }}

Please select and provide the correct option that matches the transformation shown in the image. 
Your response should end with your choice of answer.

Your response should end with your choice of answer.
\end{lstlisting}
    \caption{The prompt template used in \ref{rq:visual_input}. \textcolor{magenta}{\small \textbf{\texttt{UPLOADED IMAGE}}} is an image consists of three or more examples like in Figure~\ref{fig:level_examples}.}
    \label{fig:visual_prompt_template}
\end{figure*}

\section{Performance Decomposition in \ref{rq:matrix_input} and \ref{rq:visual_input}}
\label{app:perf_decomp}
Table~\ref{tab:perf_decomp_to_concept} provides a performance decomposition of text-based GPT-4, text-based o1-preview and multi-modal GPT-4o on our \coredatasetns-Test set. Because the rate limit of o1-preview, we conduct experiment on a subset of 50 instances. The result shows that o1-preview does not show superior results compared to the other two LLMs.

\begin{table*}[h!]
    \small
    \centering
    \begin{adjustbox}{width=\linewidth}
    \begin{tabular}{l ccccccccccc}
    \toprule
    \bf Concept       & GPT-4 (t)               & GPT-4o (v)              & o1 (t)& o1 (v)& Gemini2 FTE (v)        & DeepSeek R1 (t) & o3 (t)                   \\
    \midrule
gravity               & 60.0$_{\pm\text{8.2}}$  & 33.3$_{\pm\text{4.7}}$  & 50.0  &  80.0 & 63.3$_{\pm\text{0.3}}$  &  60.0          &  55.0$_{\pm\text{5.0}}$ \\
compression           & 50.0$_{\pm\text{20.4}}$ & 50.0$_{\pm\text{0.0}}$  & 0.0   &  50.0 & 50.0$_{\pm\text{0.0}}$  &   0.0          &   0.0$_{\pm\text{0.0}}$ \\
diffuse reflection of light
                      & 50.0$_{\pm\text{0.0}}$  & 33.3$_{\pm\text{11.8}}$ & 25.0  &  25.0 & 25.0$_{\pm\text{0.0}}$  &  25.0          &  25.0$_{\pm\text{0.0}}$ \\
lever                 & 0.0$_{\pm\text{0.0}}$   & 50.0$_{\pm\text{0.0}}$  & 16.7  &  66.7 & 77.8$_{\pm\text{0.9}}$  &  16.7          &   8.3$_{\pm\text{8.3}}$ \\
wave interference     & 83.3$_{\pm\text{11.8}}$ & 100.0$_{\pm\text{0.0}}$ & 100.0 & 100.0 & 91.7$_{\pm\text{2.1}}$  &  75.0          &  75.0$_{\pm\text{0.0}}$ \\
spectrum of light and optical filters
                      & 66.7$_{\pm\text{0.0}}$  & 88.9$_{\pm\text{15.7}}$ & 66.7  & 100.0 & 94.4$_{\pm\text{0.9}}$  & 100.0          & 100.0$_{\pm\text{0.0}}$ \\
surface tension       & 43.3$_{\pm\text{17.0}}$ & 50.0$_{\pm\text{8.2}}$  & 30.0  &  90.0 & 40.0$_{\pm\text{1.0}}$  &  40.0          &  40.0$_{\pm\text{0.0}}$ \\
nuclear fission       & 16.7$_{\pm\text{23.6}}$ & 100.0$_{\pm\text{0.0}}$ & 100.0 &  50.0 & 0.0$_{\pm\text{0.0}}$   &  50.0          &  50.0$_{\pm\text{0.0}}$ \\
nuclear fusion        & 0.0$_{\pm\text{0.0}}$   & 100.0$_{\pm\text{0.0}}$ & 50.0  &  50.0 & 33.3$_{\pm\text{33.3}}$ &  50.0          &  25.0$_{\pm\text{25.0}}$ \\
communicating vessels & 3.3$_{\pm\text{4.7}}$   & 3.3$_{\pm\text{4.7}}$   & 10.0  &  10.0 & 0.0$_{\pm\text{0.0}}$   &  50.0          &  45.0$_{\pm\text{5.0}}$ \\
diffraction of waves  & 83.3$_{\pm\text{23.6}}$ & 100.0$_{\pm\text{0.0}}$ & --    & 100.0 & 100.0$_{\pm\text{0.0}}$ & 100.0          & 100.0$_{\pm\text{0.0}}$ \\
reflection            & 86.7$_{\pm\text{4.7}}$  & 43.3$_{\pm\text{4.7}}$  & --    &  10.0 & 66.7$_{\pm\text{1.3}}$  &  70.0          &  70.0$_{\pm\text{0.0}}$ \\
refraction            & 20.0$_{\pm\text{8.2}}$  & 83.3$_{\pm\text{4.7}}$  & --    & 100.0 & 50.0$_{\pm\text{4.0}}$  &  40.0          &  50.0$_{\pm\text{10.0}}$ \\
light imaging         & 10.0$_{\pm\text{0.0}}$  & 0.0$_{\pm\text{0.0}}$   & --    &   0.0 & 16.7$_{\pm\text{0.3}}$  &   0.0          &   0.0$_{\pm\text{0.0}}$ \\
cut                   & 90.0$_{\pm\text{0.0}}$  & 73.3$_{\pm\text{4.7}}$  & --    &  60.0 & 93.3$_{\pm\text{0.3}}$  & 100.0          & 100.0$_{\pm\text{0.0}}$ \\
laser                 & 46.7$_{\pm\text{12.5}}$ & 53.3$_{\pm\text{4.7}}$  & --    &  50.0 & 26.7$_{\pm\text{2.3}}$  &  10.0          &  15.0$_{\pm\text{5.0}}$ \\
move                  & 96.7$_{\pm\text{4.7}}$  & 86.7$_{\pm\text{4.7}}$  & --    &  30.0 & 83.3$_{\pm\text{4.3}}$  &  60.0          &  70.0$_{\pm\text{10.0}}$ \\
buoyancy              & 43.3$_{\pm\text{12.5}}$ & 100.0$_{\pm\text{0.0}}$ & --    & 100.0 & 46.7$_{\pm\text{2.3}}$  &  40.0          &  40.0$_{\pm\text{0.0}}$ \\
acceleration          & 10.0$_{\pm\text{8.2}}$  & 73.3$_{\pm\text{12.5}}$ & --    &  20.0 & 46.7$_{\pm\text{0.3}}$  &  40.0          &  30.0$_{\pm\text{10.0}}$ \\
inertia               & 80.0$_{\pm\text{8.2}}$  & 6.7$_{\pm\text{4.7}}$   & --    &  10.0 & 36.7$_{\pm\text{2.3}}$  &  30.0          &  45.0$_{\pm\text{15.0}}$ \\
electricity           & 16.7$_{\pm\text{4.7}}$  & 53.3$_{\pm\text{9.4}}$  & --    &  60.0 & 30.0$_{\pm\text{0.0}}$  &  60.0          &  60.0$_{\pm\text{0.0}}$ \\
reference frame       & 27.8$_{\pm\text{3.9}}$  & 13.9$_{\pm\text{3.9}}$  & --    &  66.7 & 47.2$_{\pm\text{1.6}}$  &  25.0          &  29.1$_{\pm\text{4.1}}$ \\
repulsive force       & 20.8$_{\pm\text{5.9}}$  & 20.8$_{\pm\text{11.8}}$ & --    &  50.0 & 20.8$_{\pm\text{0.5}}$  & 100.0          &  87.5$_{\pm\text{12.5}}$ \\
diffusion             & 8.3$_{\pm\text{11.8}}$  & 100.0$_{\pm\text{0.0}}$ & --    &   0.0 & 83.3$_{\pm\text{2.1}}$  &   0.0          &   0.0$_{\pm\text{0.0}}$ \\
vortex                & 0.0$_{\pm\text{0.0}}$   & 100.0$_{\pm\text{0.0}}$ & --    &  75.0 & 91.7$_{\pm\text{2.1}}$  &   0.0          &  12.5$_{\pm\text{12.5}}$ \\
expansion             & 50.0$_{\pm\text{0.0}}$  & 75.0$_{\pm\text{0.0}}$  & --    &  75.0 & 91.7$_{\pm\text{2.1}}$  &  75.0          &  87.5$_{\pm\text{12.5}}$ \\
wave                  & 16.7$_{\pm\text{15.6}}$ & 33.3$_{\pm\text{5.9}}$  & --    &  62.5 & 25.0$_{\pm\text{0.0}}$  &  25.0          &  18.8$_{\pm\text{6.2}}$ \\
    \bottomrule
    \end{tabular}
    \end{adjustbox}
    \caption{Performance decomposition to concepts on \coredatasetns-Dev. \emph{t} and \emph{v} refer to LLMs with textual or visual inputs.}
    \label{tab:perf_decomp_to_concept}
\end{table*}

\section{Construction of Synthetic Training Data Used in \ref{rq:format_analysis}}
\label{app:synthetic_data}
We investigate whether fine-tuning LLMs on matrix property-related questions could improve their performances on our tasks. Specifically, we generate 3000 extra input-output grid pairs calculate the size, transpose, and locations of the subgrid's corner elements for these matrices as ground truths. Furthermore, since correctly recognizing the location of the subgrid may contribute more to finish the Move and Copy tasks compared to other properties, we create additional ground truths only with the gold locations of the subgrid's corner elements.

\section{Hyperparameters of Supervised Fine-Tuning in \ref{rq:format_analysis} and \ref{rq:sup_training}}
\label{app:sft_details}

For all the fine-tuning experiments, we use LoRA~\cite{hu2021lora}. We fine-tune each model for 3 epochs with a batch size of 4 on a single machine with 8 A100 GPUs. The dimension of LoRA's attention layer is set to 64, and the $\alpha$ and dropout rates are set to 16 and 0.1, respectively. The learning rate and weight decay are set to 2e-4 and 0.001, respectively.
The hyperparameters are selected according to the development performance on the synthetic matrix data in 
Appendix \ref{app:synthetic_data}.

\end{document}